\pdfoutput=1

\documentclass[11pt]{article}
\usepackage[table]{xcolor}

\usepackage{ACL2023}

\usepackage{times}
\usepackage{latexsym}
\usepackage{todonotes}
\usepackage{tabularx}
\usepackage{booktabs}       
\usepackage{graphicx}
\usepackage{pifont}
\usepackage{siunitx}
\usepackage{tikz}
\usepackage{adjustbox}
\usepackage{subcaption}
\usepackage{xcolor}
\usepackage{makecell}
\usepackage{amssymb}
\usepackage{amsmath}
\usepackage{enumitem}
\usepackage[nameinlink,noabbrev,capitalise,sort]{cleveref}
\usepackage{multirow}
\usepackage{makecell}
\usepackage{mdframed}

\usepackage[T1]{fontenc}

\usepackage[utf8]{inputenc}

\usepackage{microtype}

\usepackage{inconsolata}

\definecolor{darkgreen}{rgb}{1,50,32}
\definecolor{mygreen}{rgb}{0.0, 0.5, 0.0}
\definecolor{myred}{rgb}{0.8, 0.0, 0.0}
\definecolor{snowflake}{rgb}{0.26,0.45,0.72}
\definecolor{lightmagenta}{rgb}{0.95, 0.8, 1.0}
\definecolor{darkgray}{gray}{0.3}
\newcommand{\crope}{\textsc{CROPE}}

\title{\crope: Evaluating In-Context Adaptation of Vision and Language Models to Culture-Specific Concepts}

\author{
    Malvina Nikandrou \And
    {\bf Georgios Pantazopoulos} \And
    {\bf Nikolas Vitsakis} 
\AND
    {\bf Ioannis Konstas} \And
    {\bf Alessandro Suglia}\\
\AND
    \textnormal{Heriot-Watt University}\\
    \textnormal {\texttt{\{mn2002, gmp2000, nv2006, i.konstas, a.suglia\}}\texttt{@hw.ac.uk}}
}

\begin{document}
\maketitle
\begin{abstract}

As Vision and Language models (VLMs) are reaching users across the globe, assessing their cultural understanding has become a critical challenge.
In this paper, we introduce \crope{}, a visual question answering benchmark designed to probe the knowledge of culture-specific concepts and evaluate the capacity for cultural adaptation through contextual information.
This allows us to distinguish between parametric knowledge acquired during training and contextual knowledge provided during inference via visual and textual descriptions.
Our evaluation of several state-of-the-art open VLMs shows large performance disparities between culture-specific and common concepts in the parametric setting.
Moreover, experiments with contextual knowledge indicate that models struggle to effectively utilize multimodal information and bind culture-specific concepts to their depictions.
Our findings reveal limitations in the cultural understanding and adaptability of current VLMs that need to be addressed toward more culturally inclusive models.\footnote{Github repository \href{https://github.com/MalvinaNikandrou/crope/tree/main}{here}.}

\end{abstract}

\section{Introduction}

Recent Vision and Language models (VLMs) \cite{Qwen2VL, laurençon2024idefics2, li2024llava-onevision} have shown impressive performance across a variety of benchmarks \citep{li-etal-2023-pope, yu2024mm}. 
At the same time, frontier VLMs \cite{achiam2023gpt4, team2023gemini} have become widely accessible, making it crucial that these models can grasp the nuances of different cultures. 
Models lacking cultural awareness can affect global cultural diversity, as they can potentially contribute to content reinforcing beliefs, habits, or perspectives from more dominant cultures \citep{arora-etal-2023-probing, cao-etal-2023-assessing, tao2023auditing}.

Cultural concepts encompass both universal categories, such as birthdays, weddings, and funerals \citep{acharya2020towards}, as well as specific concepts that are primarily encountered within a particular community. 
In this work, we focus on culture-specific concepts and curate a dataset to answer the following research questions: \textit{How well do modern VLMs perform in recognizing culture-specific concepts, and can they adapt to these concepts by leveraging multimodal contextual information?} 

\begin{figure}[tb]
    \centering
    \includegraphics[width=\linewidth]{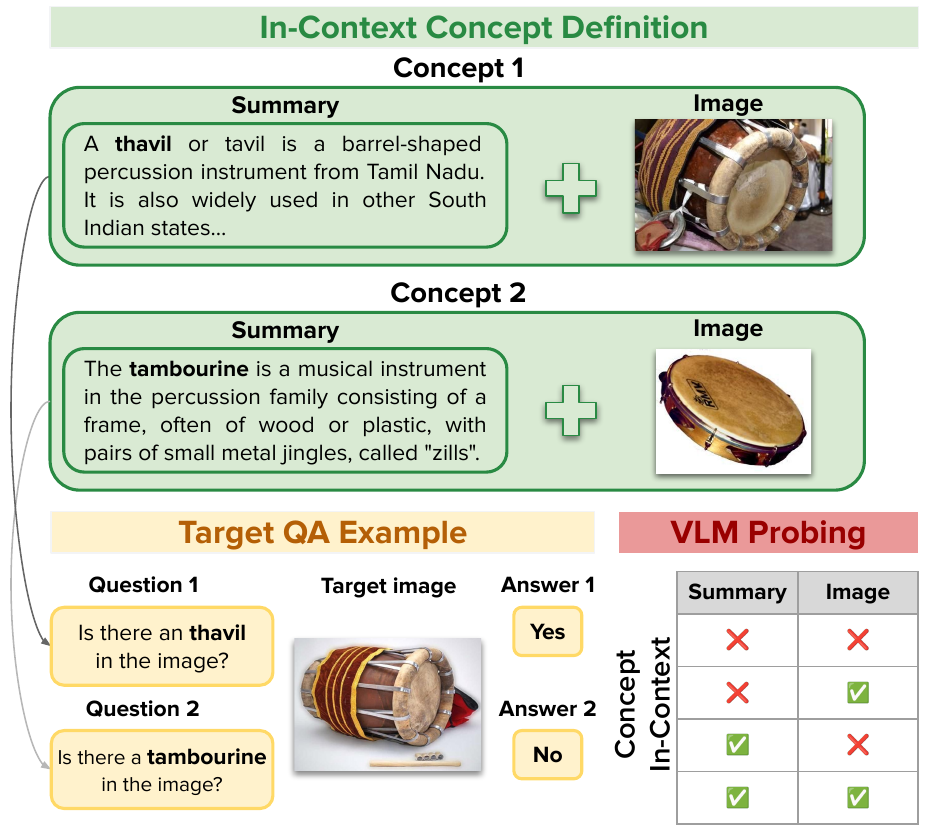}
    \caption{\crope{} probes the cultural knowledge of VLMs and assesses the effect of contextual information. Each dataset sample poses a question about the presence of a culture-specific concept within an image and is paired with demonstrative text and images that can be used as additional context to improve understanding.}
    \label{fig:into_fig}
\end{figure}

Prior studies show that Large Language Models (LLMs) \citep{johnson2022ghostmachineamericanaccent, dwivedi-etal-2023-eticor}, as well as CLIP-style vision encoders \citep{richards2024does, nwatu-etal-2023-bridging} are biased towards Western cultures \citep{liu2021visually}.
Given the paradigm of developing VLMs by combining together pre-trained vision encoders \citep{radford2021learning, zhai2023sigmoid}, and LLMs \citep{dubey2024llama, jiang2023mistral}, recent work \citep{ananthram2024see} has highlighted that these biases transfer to multimodal models.

Towards more culturally inclusive VLMs, we develop a \textit{CultuRe-specific Probing Evaluation} (\crope{}) dataset.
Similar to previous VL probing datasets \cite{hendricks-nematzadeh-2021-probing, shekhar2017foil_acl,li-etal-2023-pope}, we formulate the task as binary questions, which probe for the presence of a concept in the image, as shown in \cref{fig:into_fig}.
To stress-test a model's knowledge, we construct hard negative questions in which the concept in question and the concept in the image are visually or functionally similar. 
Although language and culture interact \citep{hovy-yang-2021-importance, hershcovich-etal-2022-challenges}, we limit our dataset to English, disentangling cultural from linguistic knowledge.

Following previous work \cite{neeman2023disentqa}, we designed \crope{} to evaluate two types of knowledge: (1) \textit{parametric}---knowledge encoded in the model weights, and (2) \textit{contextual}---external knowledge (e.g., a Wikipedia summary and corresponding image) given to the model to describe the culture-specific concept.
We experiment with several state-of-the-art open-source and open-weights VLMs with four different conditions where we vary the amount of contextual information (see \cref{fig:into_fig}). 
Our findings illustrate that with no context at all, models exhibit a considerable performance drop relative to common concepts \citep{li-etal-2023-pope} that are prevalent in most established training data for developing VLMs.
Surprisingly, when provided with contextual knowledge, the performance of most models deteriorates even more.

We analyze this behavior by inspecting the performance on an easy version of \crope, where the target concept and the concept in the question belong to separate categories (e.g., food and beverage vs animals).
In this case, most models show a performance improvement when provided with the textual information indicating that models struggle to differentiate between hard negative concepts.
Finally, we conduct a human evaluation that highlights which type of context (Wikipedia summary, image, or both) is beneficial for humans when completing the same task.
We find that the information provided by the text and the image modality is complementary for humans, which suggests that the observed model performance results from a lack of multimodal context understanding.

\section{Related Work}
\subsection{Cultural Knowledge of VLMs}

\paragraph{Evaluation of Cultural Knowledge}
Previous work has aimed to evaluate the performance of VLMs across cultures and languages. 
MaRVL \cite{liu-etal-2021-visually} tests cross-lingual transfer on visual reasoning with culturally relevant concepts,
while GD-VCR \cite{yin2021gd-vcr} focuses on commonsense reasoning regarding traditions and events from different regions, as depicted in movie scenes.
XM3600 \cite{thapliyal-etal-2022-crossmodal3600} and MaXM \cite{changpinyo2023maxm}, introduce multilingual benchmarks for image-captioning and VQA, respectively, using geographically diverse images from Open Images \cite{kuznetsova2020open}. 
However, as noted by \citet{46553}, these images do not necessarily feature culture-specific concepts despite their regional diversity.

\begin{figure*}[tb]
    \centering
    \includegraphics[width=\linewidth]{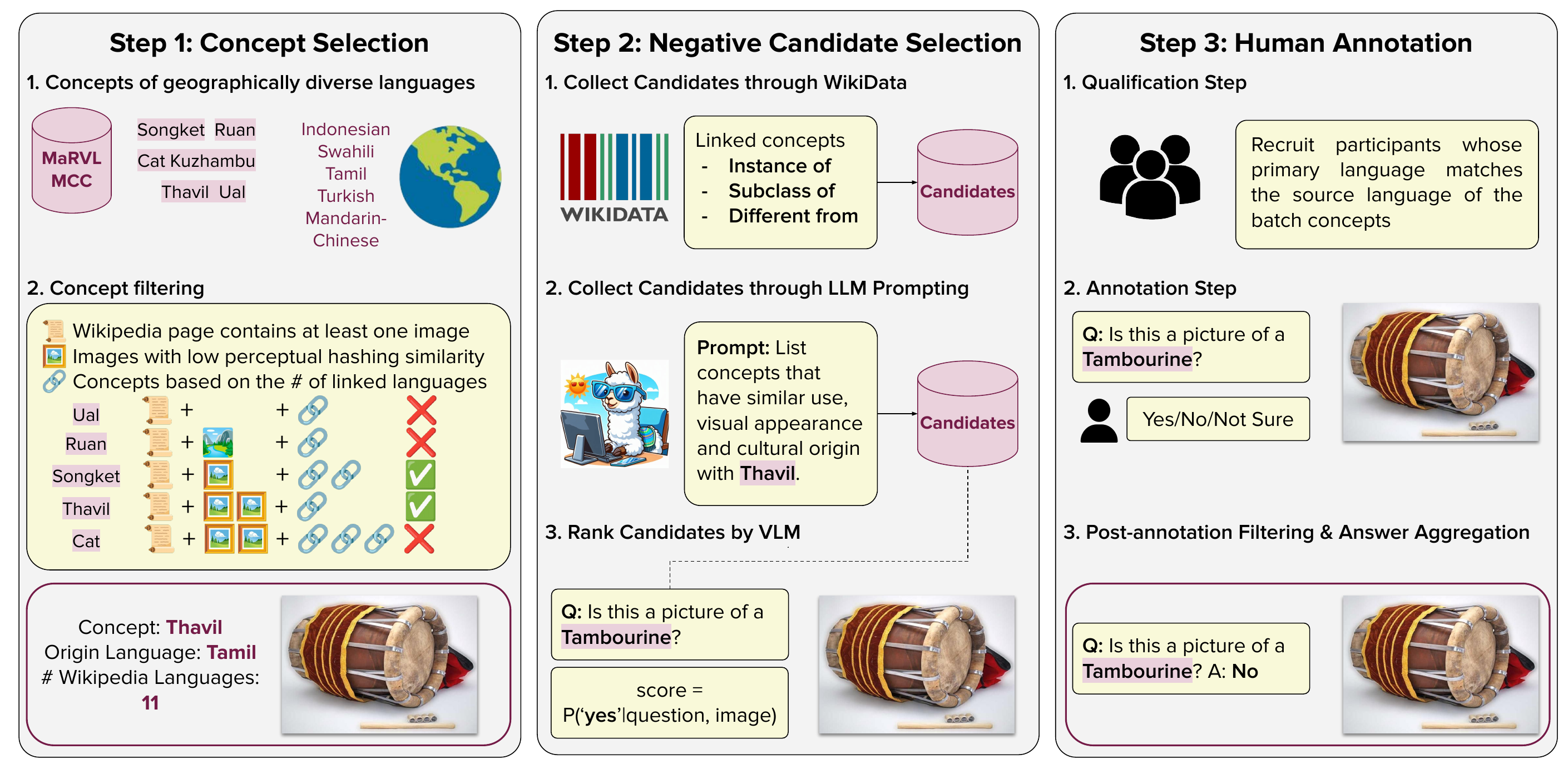}
    \caption{Overview of the dataset creation methodology. We start from a collection of concepts from geographically diverse languages. We collect a pool of challenging negative candidates from Wikidata and by prompting an LLM. Then, we use a VLM to rank candidates and sample up to three candidates per image. To verify each example, we ask human annotators who are proficient in the original concept language and English to annotate the images. Finally, we aggregate the labels and filter out ambiguous examples.}
    \label{fig:dataset}
\end{figure*}

Concurrent efforts aim to assess the capabilities of VLMs in diverse cultural contexts. These works vary in their focus, from regional traditions to multilingual capabilities.  Sea-VQA \cite{urailertprasert-etal-2024-seavqa} introduces a dataset for multi-hop reasoning on cultural concepts from eight Southeast Asian countries. GlobalRG \cite{bhatia2024local} targets geo-diverse image retrieval and visual grounding of culture-specific concepts. CulturalVQA \cite{nayak2024benchmarking} and CVQA \cite{romero2024cvqa} present knowledge-based questions centered on cultural understanding, with CVQA offering a multilingual component. 
While these works evaluate broader types of cultural knowledge, our work complements these efforts by isolating the evaluation of the recognition and adaptation to culture-specific concepts. 
CROPE additionally assesses the ability of VLMs to improve cultural understanding by leveraging non-parametric, multimodal knowledge which could serve as a scalable solution for incorporating underrepresented concepts given the constraints of model size.

\paragraph{Cultural Adaptation Methods}

Relatively few studies have focused on adapting VLMs to culture-specific concepts.
Most prior work has concentrated on encoder-only VLMs proposing approaches such as geo-diverse pretext objectives \cite{yin2023givl}, data augmentations through code-switching and image editing \cite{li-zhang-2023-mcc}, or interventions in the pretraining data composition \cite{pouget2024no-filter, ignat-etal-2024-annotations-budget}. 
In this work, we explore whether multimodal contextual knowledge provided to generative VLMs at inference can enhance the understanding of cultural concepts.

\subsection{Multimodal Context in VLMs}
Inspired by the in-context learning capabilities of LLMs, modern VLMs \cite{alayrac2022flamingo, huang2023language, laurenccon2024obelics, mckinzie2024mm1} have evolved from accepting single-image text pairs to more flexible interleaved image-text inputs.
This behavior is driven by training on multimodal web data, which enables VLMs to reason over multimodal documents, compare groups of images, or handle co-references across multimodal contexts \cite{laurençon2024idefics2, Qwen2VL, lin2024vila, xue2024xgen}.


To evaluate these capabilities, several recent benchmarks have been proposed.
These aim to evaluate the perceptual abilities \cite{wu2024qbench, fu2024blink}, cross-image reasoning \cite{li2024seed, jiang2024mantis, li2024finetuning}, or long-context processing \cite{song2024milebench} in the presence of distractor images \cite{sharma2024losing, wang2024needle}.
Contrary to the existing benchmarks, our work focuses on culture-specific concepts and aims to expand the inclusivity of VLMs through parametric or contextual knowledge.

\section{\crope{} Dataset}
The objective of \crope{} is to serve as a challenging evaluation set that probes the capabilities of modern VLMs to recognize and adapt to culture-specific concepts. \cref{fig:dataset} outlines the steps of the dataset development: 1) concept selection from different cultures, 2) negative candidate selection via model-based sampling, and 3) human annotation that verifies the correctness of each example.

\subsection{Dataset Creation Methodology}
\paragraph{Concept Selection}
We use concepts and images collected from the multilingual visual reasoning dataset MaRVL \citep{liu2021visually}, and the follow-up Multimodal Cultural Concepts (MCC) dataset \cite{li-zhang-2023-mcc}.
Both datasets contain concepts and associated images from five different language origins: Indonesian, Swahili, Tamil, Turkish, and Mandarin Chinese.
During the collection of these resources, native speakers were asked to provide concepts that are representative of the speaker's culture but common in their everyday experience.
As a result, these concepts are not necessarily unique to a particular culture \citep{cao2024exploring}, but span both universal and culture-specific concepts \citep{karamolegkou-etal-2024-vision}.

We keep the concepts for which we can recover an English Wikipedia page. 
Using the Wikipedia API, we retrieve the page summary, available images, and their captions.
We discard concepts whose Wikipedia page does not contain any images and filter the dataset images based on perceptual hashing similarity\footnote{Using the python library \href{https://github.com/JohannesBuchner/imagehash}{imagehash}.}.
To focus on culture-specific concepts, we use the number of linked Wikipedia pages in different languages as a proxy. Prior work has identified that cultural content is covered in significantly fewer languages compared to general topics \cite{10.3389/fphy.2018.00054}.
For example, the page for `Cat' is available in 267 languages, while for `Thavil' it appears only in 11 languages.
We keep the 40 concepts per language with the least number of linked Wikipedia pages (see \cref{sec:appendix-perplexity} for details).
Lastly, as Wikipedia's concept coverage varies per language, we remove concepts that appear in few languages but are well-represented in image-text datasets.

\paragraph{Negative Candidate Selection}
We aim to create a pool of negative concepts that stress-test the models' knowledge of a concept.
For this purpose, we collect negative concepts from two sources.
First, for each concept, we use the Wikidata API\footnote{\href{https://www.wikidata.org/wiki/Wikidata:REST\_API}{https://www.wikidata.org/wiki/Wikidata:REST\_API}} to collect concepts that are linked to the target concept either with the `different from' property or are children of a concept identified by the properties `subclass' and `instance of'.
For example, `Thavil' is a subclass of `Membranophones' from which we retrieve all other subclass musical instruments as possible candidates, such as the `Tambourine'.
Second, we prompt LLama3 \citep{dubey2024llama} to provide a list of 10 concepts that have similar use, visual appearance, and cultural origin with the target object.
This process creates a pool of negative candidates for each target concept.
Finally, to select challenging concepts, we rank candidates using Paligemma
\citep{beyer2024paligemma}. 
We provide the image of the positive concept, the question `Is a \texttt{<negative\_concept>}?' and measure the probability that the model answers incorrectly.
While generating the dataset, we sample up to three negative candidates based on their scores.

\paragraph{Human Annotation}
We collect ground truth answers through human annotation to 1) minimize the false negatives due to the co-occurrence of multiple concepts in images (e.g., different clothing items) and 2) ensure that the target concept is distinguishable.
For each sample, the participants are provided with a definition containing the Wikipedia image and summary for a concept and asked to determine if the concept is present in the second image.
In addition to `Yes' and `No', the participants can answer `Not Sure' for cases where the definition is not sufficient to determine the answer.

We recruit participants through the Prolific platform\footnote{\href{www.prolific.com}{www.prolific.com}}.
To ensure familiarity with the concepts depicted in the target image, our pool of participants is limited to those whose primary language matches the concept's origin language and are also fluent in English.
We recruit at least 10 participants for each language through a qualification task and collect three annotations per question. 
We discard samples where at least two participants answered `Not Sure' or there is no consensus among the annotators.
Details are provided in \cref{sec:human_anntotation}.


\begin{table*}[tb]
    \centering
    \small
    \addtolength{\tabcolsep}{-0.3em}
    \renewcommand{\arraystretch}{1.2}
    \begin{tabular}{l|cc|c|ccccc}\toprule
    & & & \bf{POPE} & \multicolumn{5}{c}{\bf{\crope{}}} \\
    \bf{Model} & \bf{MI} & \bf{ML} &  \bf{F1} & \bf{F1} & \bf{Precision} & \bf{Recall} & \bf{Yes \%} & \bf{Consistency} \\\midrule
    Majority class (No) & - & - & 33.33 & 32.26 & 50.00 & 39.22   & 0 & 0 \\ \midrule
LLaVA-1.5  \citeyearpar{liu2024improved}  & $\times$ & $\times$ & 82.19 & 62.31 & 67.15 & 67.33 & 60.64 & 38.38 \\
MOLMO \citeyearpar{deitke2024molmo}  & $\times$ & $\times$ & 83.88 & 62.50 & 70.41 & 69.35 & \bf{66.73} & 43.47 \\
LLaVA-NeXT \citeyearpar{liu2024llavanext} & $\times$ & $\times$ & 85.91 & 64.46 & 67.33 & 68.15 & 55.86 & 41.43 \\
Phi-3-Vision-128K-Instruct \citeyearpar{abdin2024phi} & $\times$ & $\times$ & 84.56 & 68.94 & 70.53 & 68.98 & 31.86 & 40.51 \\
Llama-3.2-Vision-Instruct \citeyearpar{llama3.2} & $\times$ & $\times$ & 85.06 & \textbf{79.11} & \textbf{79.38} & \textbf{80.43} & 42.95 & \textbf{64.77} \\
\midrule
Paligemma \citeyearpar{beyer2024paligemma}  & $\times$ & $\checkmark$ & 85.57 & 68.89 & 70.78 & 70.97 & 47.83 & 46.45 \\
\midrule
XGen-MM-Interleaved \citeyearpar{xue2024xgen} & $\checkmark$ & $\times$ & 86.82 & 69.24 & 74.30 & 74.95 & 61.87 & 48.86 \\
Idefics2 \citeyearpar{laurenccon2024matters} & $\checkmark$ & $\times$ & 84.13 & 70.56 & 75.15 & 75.50 & 58.78 & 52.37 \\
Mantis-Idefics2 \citeyearpar{jiang2024mantis} & $\checkmark$ & $\times$ & 84.13 & 70.79 & 72.82 & 74.31 & 53.04 & 52.69 \\
VILA \citeyearpar{lin2024vila}  & $\checkmark$ & $\times$ & 82.30 & 74.92 & 77.55 & 74.13 & 28.15 & 52.07 \\
\midrule
InternLM-XComposer-2.5 \citeyearpar{zhang2024internlm} & $\checkmark$ &  +ZH & 84.90 & 64.73 & 70.46 & 70.57 & 63.60 & 44.01 \\ 
LLaVA-OneVision \citeyearpar{li2024llava-onevision}  & $\checkmark$ &  +ZH & \bf{87.66} & 70.68 & 75.20 & 76.17 & 60.76 & 53.70 \\
Qwen2-VL-Instruct \citeyearpar{wang2024qwen2} & $\checkmark$ & $\checkmark$ & 86.91 & 74.06 & 77.56 & 79.13 & 58.17 & 58.53 \\
mPLUG-Owl3 \citeyearpar{ye2024mplug} & $\checkmark$ &  +ZH & 87.03 & 74.39 & 75.19 & 77.17 & 49.50 & 56.44 \\
    \midrule
    \textcolor{darkgray}{Gemini (gemini-1.5-flash-latest Sep 2024) \citeyearpar{team2023gemini}} & \textcolor{darkgray}{$\checkmark$} & \textcolor{darkgray}{$\checkmark$} & \textcolor{darkgray}{88.20} & \textcolor{darkgray}{78.74} & \textcolor{darkgray}{80.11} & \textcolor{darkgray}{78.60} & \textcolor{darkgray}{38.59} & \textcolor{darkgray}{60.12} \\
    \textcolor{darkgray}{Gemini (gemini-pro-latest Sep 2024) \citeyearpar{team2023gemini}} & \textcolor{darkgray}{$\checkmark$} &  \textcolor{darkgray}{$\checkmark$} & \textcolor{darkgray}{88.45} & \textcolor{darkgray}{79.27} & \textcolor{darkgray}{87.35} & \textcolor{darkgray}{72.81} & \textcolor{darkgray}{30.00} & \textcolor{darkgray}{50.19} \\
    \textcolor{darkgray}{GPT-4o (gpt-4o-2024-08-06) \citeyearpar{achiam2023gpt4}} & \textcolor{darkgray}{$\checkmark$} & \textcolor{darkgray}{$\checkmark$} & \textcolor{darkgray}{88.66} & \textcolor{darkgray}{88.87} & \textcolor{darkgray}{89.02} & \textcolor{darkgray}{88.92} & \textcolor{darkgray}{37.58} & \textcolor{darkgray}{82.49}\\
    \bottomrule
    \end{tabular}
    \caption{Zero-shot performance of models on POPE (adversarial split) and \crope{}. MI: Model has been trained with interleaved image-text data. ML: Model has been trained with multilingual image-text data. +ZH: Usage of image-text data in Chinese. The performance of closed-source models is indicated in gray.
    }
    \label{tab:results_main}
\end{table*}

\subsection{Dataset Summary}
In total, we collected 1060 examples of binary questions, where each example is also accompanied by the Wikipedia summary and images of the concept in question. 
The annotations of our study show moderate to high inter-annotator agreement (Krippendorff's alpha=0.76).
Note that the answer distribution is imbalanced (`Yes': 35.3\% , `No': 64.7\%) to probe the knowledge of the target image concepts.
We provide supplementary information regarding the dataset in \cref{sec:appendix-dataset-summary}.

\section{Experimental Setup} \label{sec:model_exp}
\paragraph{Models} We experiment with a variety of open-source and open-weights generative models up to 11B parameters (see \cref{tab:hf_model_ids}) that achieve state-of-the-art performance on established VQA benchmarks \citep{hudson2018gqa, goyal2017making}.
In our study, we categorize models based on the following: 1) whether the model has been trained with \textit{multi-image data}---which we expect should benefit from context information the most, 2) whether the model is trained with \textit{multilingual image-text data}---which we expect to show better zero-shot performance on the examined concepts.

\paragraph{Experimental Setup} To disentangle the impact of parametric and contextual knowledge~\cite{neeman2023disentqa}, our study covers four different experimental conditions:
1) \textbf{Zero-shot} (\textit{parametric}), where a model is only given the target image and question;
2) \textbf{Textual context}, where the model is also given the Wikipedia summary of the concept as additional context; 
3) \textbf{Visual context}, where the model is given the Wikipedia image of the concept as well as the caption of the image\footnote{For images without a caption. we use the template: \texttt{An image of <concept>}.};
4) \textbf{Multimodal context}, where the model receives both the summary of the concept and the corresponding exemplar image from Wikipedia.
We make a distinction between the `Multimodal' and `Visual' context conditions, although the latter technically includes both an image and its caption, as the caption does not provide a definition of the concept.
To reduce the effect of prompt sensitivity \citep{salinas-morstatter-2024-butterfly}, we use three prompts for all models and report the average performance.
We apply the same evaluation setup for zero-shot results on POPE to ensure a fair comparison. 
Finally, the conditions providing images as context apply to models that accept interleaved image-text input.

\paragraph{Evaluation Metrics} Following POPE \citep{li-etal-2023-pope}, we report the F1-score, precision, recall, as well as the percentage of positive responses.
We additionally report the consistency score \cite{hudson2018gqa}, where the model receives +1 for correctly answering `Yes' and `No' questions for a given image else 0.
To ensure reproducibility, we employ greedy decoding in all experiments.

\section{Results}\label{sec:results}
\subsection{Zero-shot Performance}
\cref{tab:results_main} shows the performance of all models in the zero-shot setting.
Models score high on the POPE adversarial split that probes for the existence of common and frequently co-occurring objects in images.
With the exception of GPT-4o, performance drops substantially on \crope{}, which targets more culturally specific objects.
Even though Gemini-1.5-Pro achieves the second-highest F1 score, there is still a considerable 9-point gap.
As the behavior of proprietary models is difficult to explain and often not reproducible, our remaining analysis focuses on open-source and open-weight models.

We observe that the F1 score of several open models (LLaVA-1.5, LLaVA-NeXT, MOLMO, InternLM-XComposer) drops by up to 20 points when they are evaluated on culture-specific concepts. 
The high Yes\% for these models indicates that they struggle to differentiate the negative candidates from the actual concept in the images.
The model with the strongest zero-shot performance is Llama-3.2, outperforming others by a large margin in terms of F1 and Consistency.
The advantage of Llama-3.2 could be explained by its extensive pretraining on a dataset of 6B image-text pairs that underwent thorough preprocessing and deduplication to maximize data diversity \cite{dubey2024llama}.
Among the four models with the highest Consistency, three remaining are trained with multilingual image-text data.
These results are in line with recent work \citep{pouget2024no-filter} that advocates for including multilingual data in the pretraining mixture, as this can enable maintaining performance on standard English benchmarks while enhancing cultural knowledge.

\begin{figure}[tb]
    \centering
    \includegraphics[width=\linewidth]{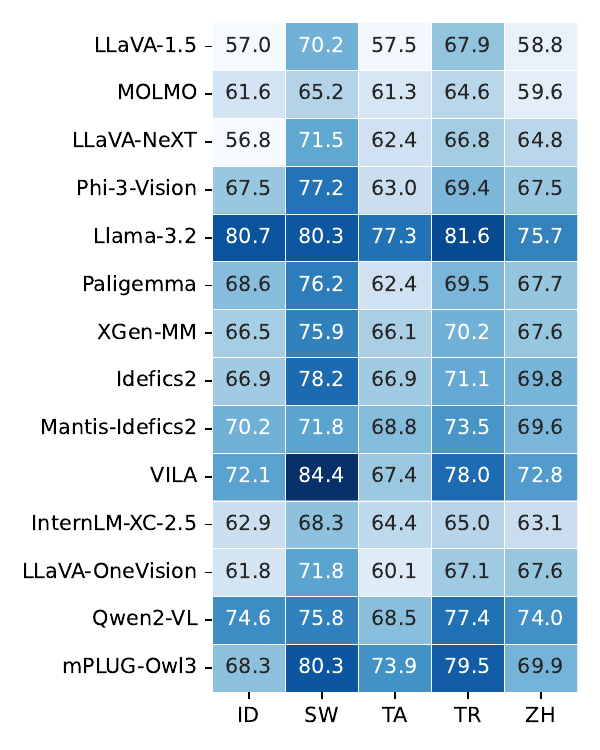}
    \caption{Zero-shot F1 score per source language.}
    \label{fig:results_per_language}
\end{figure}
\cref{fig:results_per_language} reports model performance based on the source language of the concepts.
Models perform reasonably well on images with concepts sourced from Swahili and Turkish but underperform on concepts from other high (Chinese) or mid-resource\footnote{Following the categorization of \cite{joshi-etal-2020-state}.} (Indonesian, Tamil) languages.
It is important to consider that VLMs are built by integrating vision and language backbones through joint training stages with image-text examples.
Therefore, the resource characterization of languages supported by multilingual LLMs may not accurately represent the VLM landscape.
Building on \citeauthor{pouget2024no-filter}, we need to systematically analyze the data availability and coverage of cultural concepts across different training stages of generative VLMs.

\begin{figure*}[tb]
    \centering
    \includegraphics[width=\linewidth]{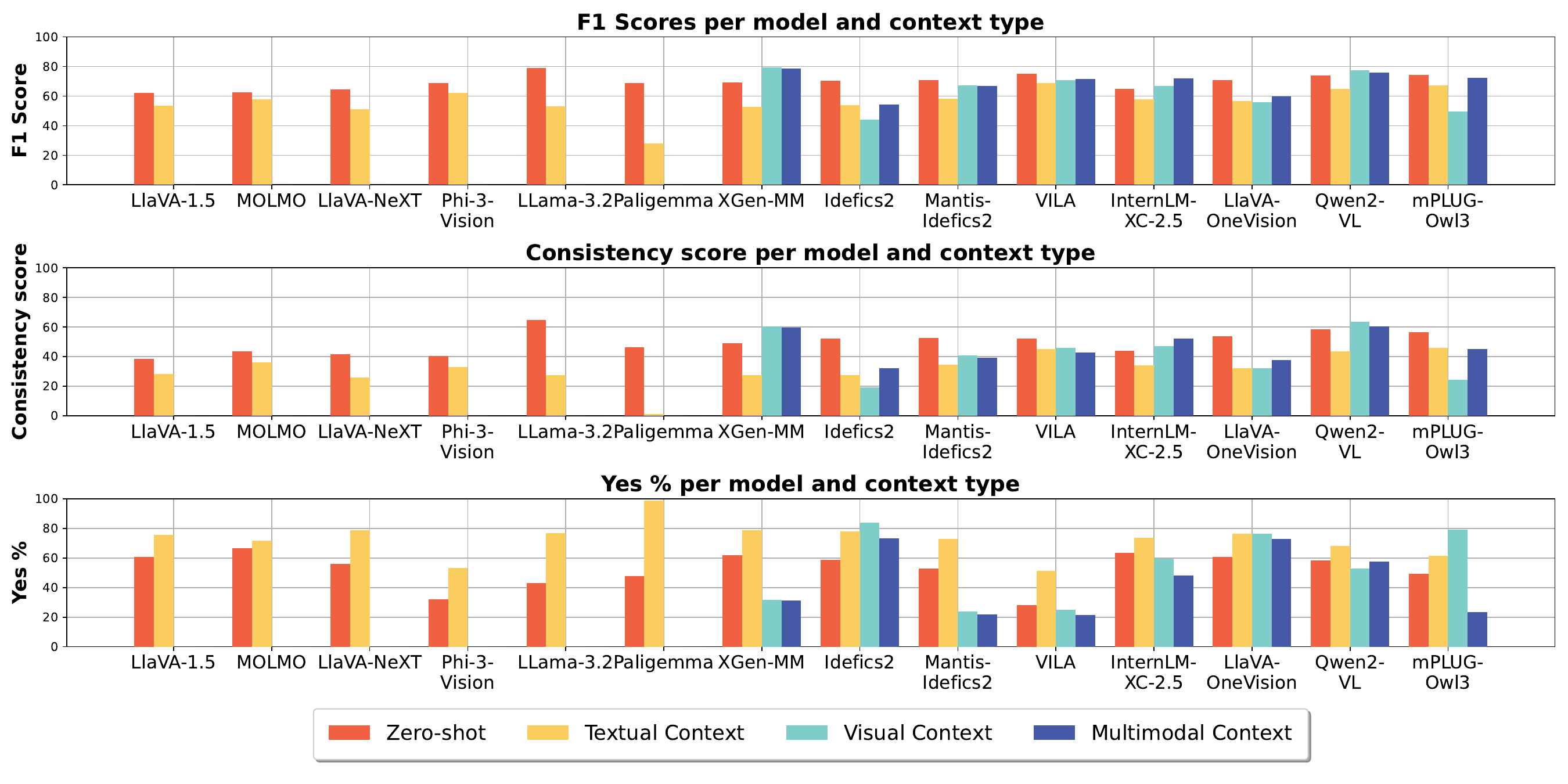}
    \caption{Performance with different context types. All VLMs are negatively impacted when including the concept summary in question (Textual Context). Out of the 7 VLMs that accept multimodal context, only XGEN-MM and InternLM-XComposer benefit from multimodal contextual information.}
    \label{fig:f1_yes_context}
\end{figure*}

\subsection{Performance with Contextual Knowledge}
We explore if the performance gap between common and culture-specific concepts can be addressed through contextual knowledge.
\cref{fig:f1_yes_context} shows the performance of all models under the different context conditions.
While contextual knowledge does not yield notable improvements in any condition, we find that the visual context is more beneficial than the textual for most models, and the multimodal context tends to be the most helpful.
We observe that in the Textual context condition, where a concept summary is provided in the prompt, all models show an increase in the percentage of `Yes' responses, leading to increased false positives.
Only three models, XGen-MMm InternLM-XComposer, and Qwen2-VL, exhibit improved performance compared to the zero-shot setting.
Nevertheless, the best-performing model-context combination does not surpass the highest zero-shot performance for open-weights VLMs.

Our findings suggest that current VLMs struggle to process multimodal contextual information.
We consider two possible reasons for this behavior.
First, the models might be sensitive to the task structure, which, due to the concept's summary, includes a relatively lengthy prompt.
To test this, we evaluate the same models on an easier version of \crope{}, which does not necessitate reasoning about subtle differences (see next paragraph).
Second, the contextual information may not suffice to disambiguate the concept in question and in the image. 
We address this by comparing against human performance in ablated contexts in \cref{sec:human_eval}.

\begin{figure}[tb]
    \centering
    \includegraphics[width=\linewidth]{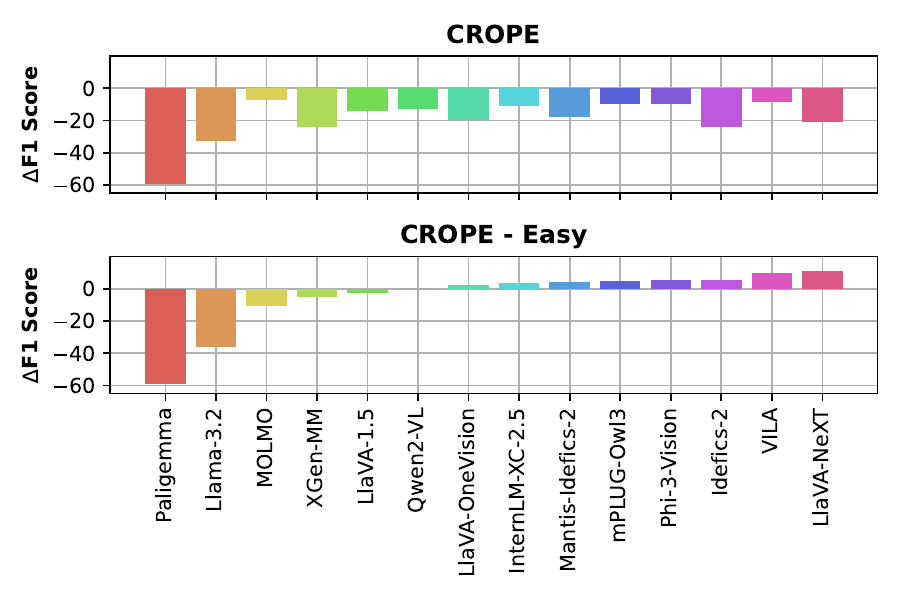}
    \caption{Relative performance of Zero-shot vs Textual conditions for the original (top) and easy (bottom) versions of \crope{}. Textual summaries benefit most models when differentiating between easier candidates.}
    \label{fig:hard_vs_easy}
\end{figure}

\paragraph{Performance with easy negative candidates}
To test the sensitivity of VLMs with regard to the input context, we create an easier version of the dataset by selecting random negative candidates.
In particular, we use the class of the target concept to sample negative candidates belonging to a different concept category.
This process creates a significantly easier version where the target image depicts a different concept category than the concept in question (e.g., a beverage vs. an animal).

We evaluate models both in zero-shot and Textual conditions. 
\cref{fig:hard_vs_easy} shows the relative performance change between the two conditions for the original and the easy dataset.
These results indicate two groupings based on the models' behavior when the length of the input prompt is increased by including the Wikipedia summary.
The first group comprises most models that show improved performance or a marginal drop in the easy version.
These models seem to be robust to the increased input prompt but struggle to differentiate between similar concepts in \crope{}.
The second group (Paligemma, Llama-3.2, MOLMO) includes models with comparable relative drops in both datasets. This behavior can be attributed to sensitivity to the longer input resulting from adding the summary.

\begin{figure}[tb]
    \centering
    \includegraphics[width=\linewidth]{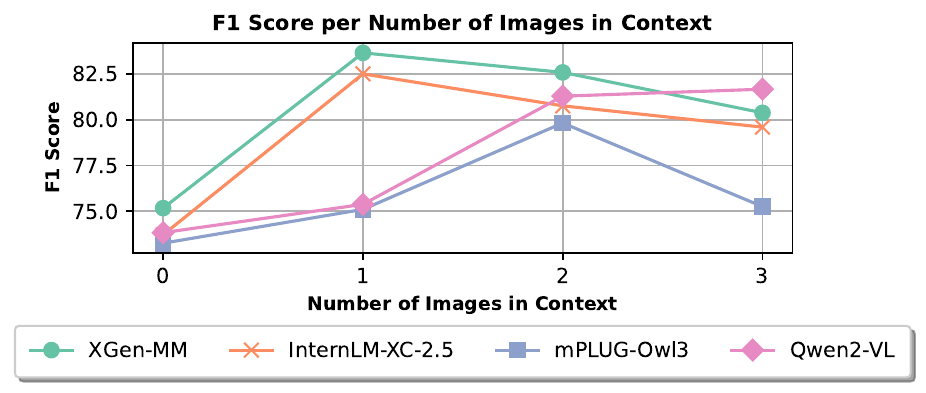}
    \caption{F1 score with varying number of images in the context.}
    \label{fig:varying_context}
\end{figure}

\paragraph{Performance with increased visual context}
We also examine the impact of providing more exemplar images in the context of the VLMs.
To do so, we keep only the samples with at least three available images and focus on the models that show the strongest performance with multimodal context.
As shown in \cref{fig:varying_context}, increasing the context images from one to two leads to a better F1 score only for mPLUG-OWL3 and Qwen2-VL but has a negative effect on the other models.
However, further increasing the context to three images hurts performance for all models except Qwen2-VL.
These results align with concurrent studies showing that the performance of highly capable VLMs deteriorates with increased images in the context \citep{zong2024vl, wang2024needle}. 

\section{Human Evaluation}\label{sec:human_eval}
We conduct a human evaluation across three conditions mirroring those of \cref{sec:model_exp}:  \textbf{Textual context}, \textbf{Visual context}, and \textbf{Multimodal context}. Each described condition relates to the modality by which information is given to participants before their annotation of examples containing unfamiliar concepts. 
Note that we do not target participants from the cultures in our dataset who would be familiar with the image concepts.
The aim is not to establish a human baseline for the zero-shot condition, as VLMs are expected to serve users from diverse backgrounds. 
Instead, we examine how the modality of information influences human judgments to put into perspective the behavior of VLMs under similar contextual settings. 


\paragraph{Participants \& Stimuli} 

We sample 100 examples with a balanced answer distribution and used power analysis \citep{cohen2013statistical, Daniels_power} to ensure that our experiment is sufficiently powered (80\%).
The design of the study is between-subjects, and we recruit 36 participants per condition, each of whom is given 10 examples.
To gauge the overall levels of familiarity of our sample, we asked participants to rate their familiarity with target concepts on a 5-point Likert scale. 
The median
of the ratings across all concepts and conditions is 1 while the 75th percentile is 2, validating that most participants were unfamiliar with the sampled concepts.

\paragraph{Experimental Design}
We use a mixed-effect regression model, with our dependent variable being the participants' binary response and our predictors matching the three conditions. 
Since participants annotated multiple examples \citep{Schielzeth_mixed_effects, raudenbush1994random}, the annotator ids were included as random factors.
Finally, we evaluate the possible effects of multiple comparisons via a Benjamini-Hochberg correction \citep{thissen2002quick, benjamini1995controlling}.

\begin{table}[tb]
\centering
\small
\setlength{\tabcolsep}{3pt}
\begin{tabular}{lccccc}
    \toprule
    & \bf{Estimate} & \bf{SE} & \bf{$z$} & \bf{$p$\textsubscript{val.}} & \bf{$p$\textsubscript{adj.}} 
    \\
    \midrule
    Intercept & 1.51 & 0.15 & 9.94 &  0.000 & \bfseries 0.000\\
    Textual Context & -0.45 & 0.20 & -2.26 &  0.024 & \bfseries  0.042\\
    Visual Context & -0.41 & 0.20 & -2.03 &  0.042 & \bfseries 0.042\\ 
    \bottomrule
\end{tabular}
\caption{Results of the regression model. p\textsubscript{val.}: significance of initial findings; p\textsubscript{adj.}: adjusted p\textsubscript{val.} after a Benjamini \& Hochberg correction.}
\label{tab:stat_results}
\end{table}

\begin{table}[tb]
    \centering
    \small
    \addtolength{\tabcolsep}{-0.2em}
    \renewcommand{\arraystretch}{1.2}
    \begin{tabular}{lccccc}\toprule
    \bf{Context Type} & \bf{F1} & \bf{Precision} & \bf{Recall} &  \bf{Yes \%} \\\midrule
    Textual & 73.12 &  74.99 & 73.29 & 63.14 \\
    Visual & 74.51 & 74.66 & 74.51 & 54.00 \\
    Multimodal & 80.92 & 80.95 & 80.89 & 58.29\\\bottomrule
    \end{tabular}
    \caption{Human performance per context condition.}
    \label{tab:human_context}
\end{table}

\paragraph{Results}
The results of the mixed regression model can be seen in \cref{tab:stat_results}. 
We report a significant negative effect on both the Textual and the Visual context conditions compared to our baseline (Multimodal). 
These results indicate that participants found the information presented through a combination of image and text to be significantly more helpful as expressed through more correct responses than when provided through either format alone.
This is in contrast with the behavior of VLMs, whose performance decays or, at best, improves minimally with the addition of any form of contextual knowledge.
Additionally, \cref{tab:human_context} shows human performance, which is well above random chance, even in the Textual condition. 
This validates that the summaries and exemplars can help reach a correct answer for unfamiliar concepts.


\section{Conclusion}
In this work, we introduce \crope{}, an evaluation benchmark for probing the parametric and contextual knowledge of VLMs on culture-specific concepts.
Our results identify a significant performance disparity of state-of-the-art open VLMs on concepts that appear commonly in VL datasets and \crope{}.
We also explore whether VLMs can adapt to culture-specific concepts with multimodal contextual information and find that most models fail to utilize this context. 
We show that this is not necessarily the case when the models are required to compare semantically distant concepts, which indicates that current VLMs struggle to reason about nuanced differences.
Conversely, our findings suggest that humans unfamiliar with the concepts in question benefit from multimodal information.

\paragraph{Discussion}
Our investigation raises the question: \textit{Are modern VLMs truly capable of learning new concepts in-context?} 
Early work \citep{tsimpoukelli2021multimodal} showed promise for VLMs capable of fast-mapping, which refers to learning to bind new concepts to images with limited contextual information \citep{carey1978acquiring}.
The authors speculate that the binding capacity of a model can be improved with richer visual or textual support.
Our analysis shows that current VLMs have not yet met this expectation, as they exhibit, at best, marginal improvements with relevant context.
Thus, processing arbitrary interleaved image-text formats remains a challenge.

Finally, we do not take the position that models should solely rely on non-parametric knowledge to perform well on tasks that require cultural understanding.
Given that benchmarks often become quickly saturated \cite{kiela-etal-2021-dynabench} and the growing interest in more pluralistic representation in VLMs, we anticipate future model iterations to `solve' the task in a zero-shot manner.
Nevertheless, \crope{} provides the opportunity to stress-test the current knowledge of culture-specific concepts embedded into the model weights, as well as the utilization of contextual knowledge.
We find that there is plenty of room for improvement across both evaluation axes, and hope that \crope{} can contribute towards the development of more capable and culturally inclusive VLMs.

\section{Limitations}
We build on previous collections of culturally relevant concepts \citep{liu2021visually} that cover only a small percentage of global cultures.
It is important that future work expands this collection to incorporate a broader range of cultural contexts for a more comprehensive evaluation.
Moreover, our dataset is limited to English. 
This does not address any potential disparities in performance across languages \cite{zhang2023dont} or cases where there is no equivalent translation \cite{majid2015semantic}.

Additionally, our findings about the behavior of VLMs are based on models with up to 11B parameters given relevant information retrieved directly from Wikipedia.
We do not examine the impact of scale or address the issue of how to retrieve informative multimodal contexts. 
This constitutes an active area of research \citep{wei2023uniir}, as it is an important consideration for practical applications. 

\section{Ethics Statement}
\crope{} contains human labels over questions regarding images that depict culture-specific concepts.
The data collection has been approved by the ethics and data protection committees of our institution.
Prior to completing the task, participants were given an information sheet detailing the purpose of the study, their rights as participants, the compensation provided for participation, and a statement assuring that no personally identifiable information would be included in any report resulting from the study.
Each participant then had the option to sign a consent form acknowledging the information provided.

For the data collection, we engaged with participants who were native speakers of the concepts' source languages to ensure inclusivity and a high degree of familiarity with the target concepts during the data collection process.
However, recruiting annotators solely based on their native language potentially overlooks minority communities.
While we acknowledge the limitations of our current work, we are hopeful that it will contribute to advancing cultural understanding of modern VLMs. 
We encourage future research to explore additional criteria within cultural groups to facilitate more representative sampling.

\section*{Acknowledgements}
This work was supported by the Edinburgh International Data Facility (EIDF) and the Data-Driven Innovation Program at the University of Edinburgh.
In addition, the authors acknowledge the use of the HWU high-performance computing facility (DMOG) and associated support services in the completion of this work.

\bibliography{acl2023}

\begin{thebibliography}{91}
\expandafter\ifx\csname natexlab\endcsname\relax\def\natexlab#1{#1}\fi

\bibitem[{Abdin et~al.(2024)Abdin, Jacobs, Awan, Aneja, Awadallah, Awadalla, Bach, Bahree, Bakhtiari, Behl et~al.}]{abdin2024phi}
Marah Abdin, Sam~Ade Jacobs, Ammar~Ahmad Awan, Jyoti Aneja, Ahmed Awadallah, Hany Awadalla, Nguyen Bach, Amit Bahree, Arash Bakhtiari, Harkirat Behl, et~al. 2024.
\newblock \href {https://arxiv.org/abs/2404.14219} {Phi-3 technical report: A highly capable language model locally on your phone}.
\newblock \emph{arXiv preprint arXiv:2404.14219}.

\bibitem[{Acharya et~al.(2020)Acharya, Talamadupula, and Finlayson}]{acharya2020towards}
Anurag Acharya, Kartik Talamadupula, and Mark~A Finlayson. 2020.
\newblock \href {https://usc-isi-i2.github.io/AAAI21workshop/papers/Acharya_CSKGsAAAI-21.pdf} {Towards an atlas of cultural commonsense for machine reasoning}.
\newblock In \emph{Workshop on Common Sense Knowledge Graphs (CSKGs)}.

\bibitem[{Achiam et~al.(2023)Achiam, Adler, Agarwal, Ahmad, Akkaya, Aleman, Almeida, Altenschmidt, Altman, Anadkat et~al.}]{achiam2023gpt4}
Josh Achiam, Steven Adler, Sandhini Agarwal, Lama Ahmad, Ilge Akkaya, Florencia~Leoni Aleman, Diogo Almeida, Janko Altenschmidt, Sam Altman, Shyamal Anadkat, et~al. 2023.
\newblock \href {https://arxiv.org/abs/2303.08774} {Gpt-4 technical report}.
\newblock \emph{arXiv preprint arXiv:2303.08774}.

\bibitem[{Alayrac et~al.(2022)Alayrac, Donahue, Luc, Miech, Barr, Hasson, Lenc, Mensch, Millican, Reynolds, Ring, Rutherford, Cabi, Han, Gong, Samangooei, Monteiro, Menick, Borgeaud, Brock, Nematzadeh, Sharifzadeh, Binkowski, Barreira, Vinyals, Zisserman, and Simonyan}]{alayrac2022flamingo}
Jean-Baptiste Alayrac, Jeff Donahue, Pauline Luc, Antoine Miech, Iain Barr, Yana Hasson, Karel Lenc, Arthur Mensch, Katherine Millican, Malcolm Reynolds, Roman Ring, Eliza Rutherford, Serkan Cabi, Tengda Han, Zhitao Gong, Sina Samangooei, Marianne Monteiro, Jacob Menick, Sebastian Borgeaud, Andrew Brock, Aida Nematzadeh, Sahand Sharifzadeh, Mikolaj Binkowski, Ricardo Barreira, Oriol Vinyals, Andrew Zisserman, and Karen Simonyan. 2022.
\newblock \href {https://openreview.net/forum?id=EbMuimAbPbs} {Flamingo: a visual language model for few-shot learning}.
\newblock In \emph{Advances in Neural Information Processing Systems}.

\bibitem[{Ananthram et~al.(2024)Ananthram, Stengel-Eskin, Vondrick, Bansal, and McKeown}]{ananthram2024see}
Amith Ananthram, Elias Stengel-Eskin, Carl Vondrick, Mohit Bansal, and Kathleen McKeown. 2024.
\newblock \href {https://arxiv.org/abs/2406.11665} {See it from my perspective: Diagnosing the western cultural bias of large vision-language models in image understanding}.
\newblock \emph{arXiv preprint arXiv:2406.11665}.

\bibitem[{Arora et~al.(2023)Arora, Kaffee, and Augenstein}]{arora-etal-2023-probing}
Arnav Arora, Lucie-aim{\'e}e Kaffee, and Isabelle Augenstein. 2023.
\newblock \href {https://doi.org/10.18653/v1/2023.c3nlp-1.12} {Probing pre-trained language models for cross-cultural differences in values}.
\newblock In \emph{Proceedings of the First Workshop on Cross-Cultural Considerations in NLP (C3NLP)}, pages 114--130. Association for Computational Linguistics.

\bibitem[{Benjamini and Hochberg(1995)}]{benjamini1995controlling}
Yoav Benjamini and Yosef Hochberg. 1995.
\newblock \href {https://rss.onlinelibrary.wiley.com/doi/10.1111/j.2517-6161.1995.tb02031.x} {Controlling the false discovery rate: a practical and powerful approach to multiple testing}.
\newblock \emph{Journal of the Royal statistical society: series B (Methodological)}, 57(1):289--300.

\bibitem[{Beyer et~al.(2024)Beyer, Steiner, Pinto, Kolesnikov, Wang, Salz, Neumann, Alabdulmohsin, Tschannen, Bugliarello et~al.}]{beyer2024paligemma}
Lucas Beyer, Andreas Steiner, Andr{\'e}~Susano Pinto, Alexander Kolesnikov, Xiao Wang, Daniel Salz, Maxim Neumann, Ibrahim Alabdulmohsin, Michael Tschannen, Emanuele Bugliarello, et~al. 2024.
\newblock \href {https://arxiv.org/abs/2407.07726} {Paligemma: A versatile 3b vlm for transfer}.
\newblock \emph{arXiv preprint arXiv:2407.07726}.

\bibitem[{Bhatia et~al.(2024)Bhatia, Ravi, Chinchure, Hwang, and Shwartz}]{bhatia2024local}
Mehar Bhatia, Sahithya Ravi, Aditya Chinchure, Eunjeong Hwang, and Vered Shwartz. 2024.
\newblock \href {https://arxiv.org/abs/2407.00263} {From local concepts to universals: Evaluating the multicultural understanding of vision-language models}.
\newblock \emph{arXiv preprint arXiv:2407.00263}.

\bibitem[{Borin et~al.(2013)Borin, Comrie, and Saxena}]{borin2013intercontinental}
Lars Borin, Bernard Comrie, and Anju Saxena. 2013.
\newblock \href {https://www.degruyter.com/document/doi/10.1515/9783110305258.285/pdf?licenseType=restricted} {The intercontinental dictionary series--a rich and principled database for language comparison}.
\newblock \emph{Approaches to measuring linguistic differences}, 285:302.

\bibitem[{Cao et~al.(2024)Cao, Li, Li, Yuan, and Hershcovich}]{cao2024exploring}
Yong Cao, Wenyan Li, Jiaang Li, Yifei Yuan, and Daniel Hershcovich. 2024.
\newblock \href {https://arxiv.org/abs/2402.06015} {Exploring visual culture awareness in gpt-4v: A comprehensive probing}.
\newblock \emph{arXiv preprint arXiv:2402.06015}.

\bibitem[{Cao et~al.(2023)Cao, Zhou, Lee, Cabello, Chen, and Hershcovich}]{cao-etal-2023-assessing}
Yong Cao, Li~Zhou, Seolhwa Lee, Laura Cabello, Min Chen, and Daniel Hershcovich. 2023.
\newblock \href {https://doi.org/10.18653/v1/2023.c3nlp-1.7} {Assessing cross-cultural alignment between {C}hat{GPT} and human societies: An empirical study}.
\newblock In \emph{Proceedings of the First Workshop on Cross-Cultural Considerations in NLP (C3NLP)}, pages 53--67. Association for Computational Linguistics.

\bibitem[{Carey and Bartlett(1978)}]{carey1978acquiring}
Susan Carey and Elsa Bartlett. 1978.
\newblock \href {https://eric.ed.gov/?id=ED198703} {Acquiring a single new word.}
\newblock \emph{Papers and Reports on Child Language Development}.

\bibitem[{Changpinyo et~al.(2023)Changpinyo, Xue, Yarom, Thapliyal, Szpektor, Amelot, Chen, and Soricut}]{changpinyo2023maxm}
Soravit Changpinyo, Linting Xue, Michal Yarom, Ashish Thapliyal, Idan Szpektor, Julien Amelot, Xi~Chen, and Radu Soricut. 2023.
\newblock \href {https://aclanthology.org/2023.findings-emnlp.176.pdf} {Maxm: Towards multilingual visual question answering}.
\newblock In \emph{Findings of the Association for Computational Linguistics: EMNLP 2023}, pages 2667--2682.

\bibitem[{Chen et~al.(2024)Chen, Wang, Tian, Ye, Gao, Cui, Tong, Hu, Luo, Ma et~al.}]{chen2024far}
Zhe Chen, Weiyun Wang, Hao Tian, Shenglong Ye, Zhangwei Gao, Erfei Cui, Wenwen Tong, Kongzhi Hu, Jiapeng Luo, Zheng Ma, et~al. 2024.
\newblock \href {https://link.springer.com/article/10.1007/s11432-024-4231-5} {How far are we to gpt-4v? closing the gap to commercial multimodal models with open-source suites}.
\newblock \emph{Science China Information Sciences}, 67(12):220101.

\bibitem[{Cohen(2013)}]{cohen2013statistical}
Jacob Cohen. 2013.
\newblock \href {https://www.taylorfrancis.com/books/mono/10.4324/9780203771587/statistical-power-analysis-behavioral-sciences-jacob-cohen} {\emph{Statistical power analysis for the behavioral sciences}}.
\newblock routledge.

\bibitem[{Deitke et~al.(2024)Deitke, Clark, Lee, Tripathi, Yang, Park, Salehi, Muennighoff, Lo, Soldaini et~al.}]{deitke2024molmo}
Matt Deitke, Christopher Clark, Sangho Lee, Rohun Tripathi, Yue Yang, Jae~Sung Park, Mohammadreza Salehi, Niklas Muennighoff, Kyle Lo, Luca Soldaini, et~al. 2024.
\newblock \href {https://arxiv.org/abs/2409.17146} {Molmo and pixmo: Open weights and open data for state-of-the-art multimodal models}.

\bibitem[{Dubey et~al.(2024)Dubey, Jauhri, Pandey, Kadian, Al-Dahle, Letman, Mathur, Schelten, Yang, Fan et~al.}]{dubey2024llama}
Abhimanyu Dubey, Abhinav Jauhri, Abhinav Pandey, Abhishek Kadian, Ahmad Al-Dahle, Aiesha Letman, Akhil Mathur, Alan Schelten, Amy Yang, Angela Fan, et~al. 2024.
\newblock \href {https://arxiv.org/abs/2407.21783} {The llama 3 herd of models}.
\newblock \emph{arXiv preprint arXiv:2407.21783}.

\bibitem[{Dwivedi et~al.(2023)Dwivedi, Lavania, and Modi}]{dwivedi-etal-2023-eticor}
Ashutosh Dwivedi, Pradhyumna Lavania, and Ashutosh Modi. 2023.
\newblock \href {https://doi.org/10.18653/v1/2023.emnlp-main.428} {{E}ti{C}or: Corpus for analyzing {LLM}s for etiquettes}.
\newblock In \emph{Proceedings of the 2023 Conference on Empirical Methods in Natural Language Processing}, pages 6921--6931. Association for Computational Linguistics.

\bibitem[{Fu et~al.(2024)Fu, Hu, Li, Feng, Wang, Lin, Roth, Smith, Ma, and Krishna}]{fu2024blink}
Xingyu Fu, Yushi Hu, Bangzheng Li, Yu~Feng, Haoyu Wang, Xudong Lin, Dan Roth, Noah~A Smith, Wei-Chiu Ma, and Ranjay Krishna. 2024.
\newblock \href {https://arxiv.org/abs/2404.12390} {Blink: Multimodal large language models can see but not perceive}.
\newblock \emph{arXiv preprint arXiv:2404.12390}.

\bibitem[{Goyal et~al.(2017)Goyal, Khot, Summers-Stay, Batra, and Parikh}]{goyal2017making}
Yash Goyal, Tejas Khot, Douglas Summers-Stay, Dhruv Batra, and Devi Parikh. 2017.
\newblock \href {https://openaccess.thecvf.com/content_cvpr_2017/html/Goyal_Making_the_v_CVPR_2017_paper.html} {Making the v in vqa matter: Elevating the role of image understanding in visual question answering}.
\newblock In \emph{Proceedings of the IEEE conference on computer vision and pattern recognition}, pages 6904--6913.

\bibitem[{Groeneveld et~al.(2024)Groeneveld, Beltagy, Walsh, Bhagia, Kinney, Tafjord, Jha, Ivison, Magnusson, Wang, Arora, Atkinson, Authur, Chandu, Cohan, Dumas, Elazar, Gu, Hessel, Khot, Merrill, Morrison, Muennighoff, Naik, Nam, Peters, Pyatkin, Ravichander, Schwenk, Shah, Smith, Strubell, Subramani, Wortsman, Dasigi, Lambert, Richardson, Zettlemoyer, Dodge, Lo, Soldaini, Smith, and Hajishirzi}]{groeneveld-etal-2024-olmo}
Dirk Groeneveld, Iz~Beltagy, Evan Walsh, Akshita Bhagia, Rodney Kinney, Oyvind Tafjord, Ananya Jha, Hamish Ivison, Ian Magnusson, Yizhong Wang, Shane Arora, David Atkinson, Russell Authur, Khyathi Chandu, Arman Cohan, Jennifer Dumas, Yanai Elazar, Yuling Gu, Jack Hessel, Tushar Khot, William Merrill, Jacob Morrison, Niklas Muennighoff, Aakanksha Naik, Crystal Nam, Matthew Peters, Valentina Pyatkin, Abhilasha Ravichander, Dustin Schwenk, Saurabh Shah, William Smith, Emma Strubell, Nishant Subramani, Mitchell Wortsman, Pradeep Dasigi, Nathan Lambert, Kyle Richardson, Luke Zettlemoyer, Jesse Dodge, Kyle Lo, Luca Soldaini, Noah Smith, and Hannaneh Hajishirzi. 2024.
\newblock \href {https://doi.org/10.18653/v1/2024.acl-long.841} {{OLM}o: Accelerating the science of language models}.
\newblock In \emph{Proceedings of the 62nd Annual Meeting of the Association for Computational Linguistics (Volume 1: Long Papers)}, pages 15789--15809. Association for Computational Linguistics.

\bibitem[{Hendricks and Nematzadeh(2021)}]{hendricks-nematzadeh-2021-probing}
Lisa~Anne Hendricks and Aida Nematzadeh. 2021.
\newblock \href {https://doi.org/10.18653/v1/2021.findings-acl.318} {Probing image-language transformers for verb understanding}.
\newblock In \emph{Findings of the Association for Computational Linguistics: ACL-IJCNLP 2021}, pages 3635--3644, Online. Association for Computational Linguistics.

\bibitem[{Hershcovich et~al.(2022)Hershcovich, Frank, Lent, de~Lhoneux, Abdou, Brandl, Bugliarello, Cabello~Piqueras, Chalkidis, Cui, Fierro, Margatina, Rust, and S{\o}gaard}]{hershcovich-etal-2022-challenges}
Daniel Hershcovich, Stella Frank, Heather Lent, Miryam de~Lhoneux, Mostafa Abdou, Stephanie Brandl, Emanuele Bugliarello, Laura Cabello~Piqueras, Ilias Chalkidis, Ruixiang Cui, Constanza Fierro, Katerina Margatina, Phillip Rust, and Anders S{\o}gaard. 2022.
\newblock \href {https://doi.org/10.18653/v1/2022.acl-long.482} {Challenges and strategies in cross-cultural {NLP}}.
\newblock In \emph{Proceedings of the 60th Annual Meeting of the Association for Computational Linguistics (Volume 1: Long Papers)}, pages 6997--7013, Dublin, Ireland. Association for Computational Linguistics.

\bibitem[{Hovy and Yang(2021)}]{hovy-yang-2021-importance}
Dirk Hovy and Diyi Yang. 2021.
\newblock \href {https://doi.org/10.18653/v1/2021.naacl-main.49} {The importance of modeling social factors of language: Theory and practice}.
\newblock In \emph{Proceedings of the 2021 Conference of the North American Chapter of the Association for Computational Linguistics: Human Language Technologies}, pages 588--602, Online. Association for Computational Linguistics.

\bibitem[{Huang et~al.(2023)Huang, Dong, Wang, Hao, Singhal, Ma, Lv, Cui, Mohammed, Patra et~al.}]{huang2023language}
Shaohan Huang, Li~Dong, Wenhui Wang, Yaru Hao, Saksham Singhal, Shuming Ma, Tengchao Lv, Lei Cui, Owais~Khan Mohammed, Barun Patra, et~al. 2023.
\newblock \href {https://proceedings.neurips.cc/paper_files/paper/2023/hash/e425b75bac5742a008d643826428787c-Abstract-Conference.html} {Language is not all you need: Aligning perception with language models}.
\newblock \emph{Advances in Neural Information Processing Systems}, 36:72096--72109.

\bibitem[{Hudson and Manning(2019)}]{hudson2018gqa}
Drew~A Hudson and Christopher~D Manning. 2019.
\newblock \href {https://openaccess.thecvf.com/content_CVPR_2019/html/Hudson_GQA_A_New_Dataset_for_Real-World_Visual_Reasoning_and_Compositional_CVPR_2019_paper.html} {Gqa: A new dataset for real-world visual reasoning and compositional question answering}.
\newblock \emph{Conference on Computer Vision and Pattern Recognition (CVPR)}.

\bibitem[{Ignat et~al.(2024)Ignat, Bai, Nwatu, and Mihalcea}]{ignat-etal-2024-annotations-budget}
Oana Ignat, Longju Bai, Joan~C. Nwatu, and Rada Mihalcea. 2024.
\newblock \href {https://aclanthology.org/2024.lrec-main.112} {Annotations on a budget: Leveraging geo-data similarity to balance model performance and annotation cost}.
\newblock In \emph{Proceedings of the 2024 Joint International Conference on Computational Linguistics, Language Resources and Evaluation (LREC-COLING 2024)}, pages 1239--1259. ELRA and ICCL.

\bibitem[{Jiang et~al.(2023)Jiang, Sablayrolles, Mensch, Bamford, Chaplot, Casas, Bressand, Lengyel, Lample, Saulnier et~al.}]{jiang2023mistral}
Albert~Q Jiang, Alexandre Sablayrolles, Arthur Mensch, Chris Bamford, Devendra~Singh Chaplot, Diego de~las Casas, Florian Bressand, Gianna Lengyel, Guillaume Lample, Lucile Saulnier, et~al. 2023.
\newblock \href {https://arxiv.org/abs/2310.06825} {Mistral 7b}.
\newblock \emph{ArXiv}, abs/2310.06825.

\bibitem[{Jiang et~al.(2024)Jiang, He, Zeng, Wei, Ku, Liu, and Chen}]{jiang2024mantis}
Dongfu Jiang, Xuan He, Huaye Zeng, Cong Wei, Max Ku, Qian Liu, and Wenhu Chen. 2024.
\newblock \href {https://arxiv.org/abs/2405.01483} {Mantis: Interleaved multi-image instruction tuning}.
\newblock \emph{arXiv preprint arXiv:2405.01483}.

\bibitem[{Johnson et~al.(2022)Johnson, Pistilli, Menédez-González, Duran, Panai, Kalpokiene, and Bertulfo}]{johnson2022ghostmachineamericanaccent}
Rebecca~L Johnson, Giada Pistilli, Natalia Menédez-González, Leslye Denisse~Dias Duran, Enrico Panai, Julija Kalpokiene, and Donald~Jay Bertulfo. 2022.
\newblock \href {https://arxiv.org/abs/2203.07785} {The ghost in the machine has an american accent: value conflict in gpt-3}.
\newblock \emph{ArXiv}, abs/2203.07785.

\bibitem[{Joshi et~al.(2020)Joshi, Santy, Budhiraja, Bali, and Choudhury}]{joshi-etal-2020-state}
Pratik Joshi, Sebastin Santy, Amar Budhiraja, Kalika Bali, and Monojit Choudhury. 2020.
\newblock \href {https://doi.org/10.18653/v1/2020.acl-main.560} {The state and fate of linguistic diversity and inclusion in the {NLP} world}.
\newblock In \emph{Proceedings of the 58th Annual Meeting of the Association for Computational Linguistics}, pages 6282--6293, Online. Association for Computational Linguistics.

\bibitem[{Kandpal et~al.(2023)Kandpal, Deng, Roberts, Wallace, and Raffel}]{pmlr-v202-kandpal23a}
Nikhil Kandpal, Haikang Deng, Adam Roberts, Eric Wallace, and Colin Raffel. 2023.
\newblock \href {https://proceedings.mlr.press/v202/kandpal23a.html} {Large language models struggle to learn long-tail knowledge}.
\newblock In \emph{Proceedings of the 40th International Conference on Machine Learning}, volume 202 of \emph{Proceedings of Machine Learning Research}, pages 15696--15707. PMLR.

\bibitem[{Karamolegkou et~al.(2024)Karamolegkou, Rust, Cui, Cao, S{\o}gaard, and Hershcovich}]{karamolegkou-etal-2024-vision}
Antonia Karamolegkou, Phillip Rust, Ruixiang Cui, Yong Cao, Anders S{\o}gaard, and Daniel Hershcovich. 2024.
\newblock \href {https://doi.org/10.18653/v1/2024.hucllm-1.5} {Vision-language models under cultural and inclusive considerations}.
\newblock In \emph{Proceedings of the 1st Human-Centered Large Language Modeling Workshop}, pages 53--66. ACL.

\bibitem[{Kiela et~al.(2021)Kiela, Bartolo, Nie, Kaushik, Geiger, Wu, Vidgen, Prasad, Singh, Ringshia, Ma, Thrush, Riedel, Waseem, Stenetorp, Jia, Bansal, Potts, and Williams}]{kiela-etal-2021-dynabench}
Douwe Kiela, Max Bartolo, Yixin Nie, Divyansh Kaushik, Atticus Geiger, Zhengxuan Wu, Bertie Vidgen, Grusha Prasad, Amanpreet Singh, Pratik Ringshia, Zhiyi Ma, Tristan Thrush, Sebastian Riedel, Zeerak Waseem, Pontus Stenetorp, Robin Jia, Mohit Bansal, Christopher Potts, and Adina Williams. 2021.
\newblock \href {https://doi.org/10.18653/v1/2021.naacl-main.324} {Dynabench: Rethinking benchmarking in {NLP}}.
\newblock In \emph{Proceedings of the 2021 Conference of the North American Chapter of the Association for Computational Linguistics: Human Language Technologies}, pages 4110--4124, Online. Association for Computational Linguistics.

\bibitem[{Kuznetsova et~al.(2020)Kuznetsova, Rom, Alldrin, Uijlings, Krasin, Pont-Tuset, Kamali, Popov, Malloci, Kolesnikov et~al.}]{kuznetsova2020open}
Alina Kuznetsova, Hassan Rom, Neil Alldrin, Jasper Uijlings, Ivan Krasin, Jordi Pont-Tuset, Shahab Kamali, Stefan Popov, Matteo Malloci, Alexander Kolesnikov, et~al. 2020.
\newblock \href {https://link.springer.com/article/10.1007/s11263-020-01316-z} {The open images dataset v4: Unified image classification, object detection, and visual relationship detection at scale}.
\newblock \emph{International journal of computer vision}, 128(7):1956--1981.

\bibitem[{Lakens and Caldwell(2021)}]{Daniels_power}
Daniël Lakens and Aaron~R. Caldwell. 2021.
\newblock \href {https://doi.org/10.1177/2515245920951503} {Simulation-based power analysis for factorial analysis of variance designs}.
\newblock \emph{Advances in Methods and Practices in Psychological Science}, 4(1):2515245920951503.

\bibitem[{Lauren{\c{c}}on et~al.(2024{\natexlab{a}})Lauren{\c{c}}on, Saulnier, Tronchon, Bekman, Singh, Lozhkov, Wang, Karamcheti, Rush, Kiela et~al.}]{laurenccon2024obelics}
Hugo Lauren{\c{c}}on, Lucile Saulnier, L{\'e}o Tronchon, Stas Bekman, Amanpreet Singh, Anton Lozhkov, Thomas Wang, Siddharth Karamcheti, Alexander Rush, Douwe Kiela, et~al. 2024{\natexlab{a}}.
\newblock \href {https://openreview.net/forum?id=SKN2hflBIZ} {Obelics: An open web-scale filtered dataset of interleaved image-text documents}.
\newblock \emph{Advances in Neural Information Processing Systems}, 36.

\bibitem[{Lauren{\c{c}}on et~al.(2024{\natexlab{b}})Lauren{\c{c}}on, Tronchon, Cord, and Sanh}]{laurenccon2024matters}
Hugo Lauren{\c{c}}on, L{\'e}o Tronchon, Matthieu Cord, and Victor Sanh. 2024{\natexlab{b}}.
\newblock \href {https://arxiv.org/abs/2405.02246} {What matters when building vision-language models?}
\newblock \emph{arXiv preprint arXiv:2405.02246}.

\bibitem[{Laurençon et~al.(2024)Laurençon, Tronchon, Cord, and Sanh}]{laurençon2024idefics2}
Hugo Laurençon, Léo Tronchon, Matthieu Cord, and Victor Sanh. 2024.
\newblock \href {http://arxiv.org/abs/2405.02246} {What matters when building vision-language models?}

\bibitem[{Li et~al.(2024{\natexlab{a}})Li, Zhang, Guo, Zhang, Li, Zhang, Zhang, Li, Liu, and Li}]{li2024llava-onevision}
Bo~Li, Yuanhan Zhang, Dong Guo, Renrui Zhang, Feng Li, Hao Zhang, Kaichen Zhang, Yanwei Li, Ziwei Liu, and Chunyuan Li. 2024{\natexlab{a}}.
\newblock \href {https://arxiv.org/abs/2408.03326} {Llava-onevision: Easy visual task transfer}.
\newblock \emph{arXiv preprint arXiv:2408.03326}.

\bibitem[{Li et~al.(2024{\natexlab{b}})Li, Ge, Ge, Wang, Wang, Zhang, and Shan}]{li2024seed}
Bohao Li, Yuying Ge, Yixiao Ge, Guangzhi Wang, Rui Wang, Ruimao Zhang, and Ying Shan. 2024{\natexlab{b}}.
\newblock \href {https://openaccess.thecvf.com/content/CVPR2024/html/Li_SEED-Bench_Benchmarking_Multimodal_Large_Language_Models_CVPR_2024_paper.html} {Seed-bench: Benchmarking multimodal large language models}.
\newblock In \emph{Proceedings of the IEEE/CVF Conference on Computer Vision and Pattern Recognition}, pages 13299--13308.

\bibitem[{Li et~al.(2024{\natexlab{c}})Li, Pan, Ge, Gao, Ji, Zhang, Chua, Tang, Zhang, and Zhuang}]{li2024finetuning}
Juncheng Li, Kaihang Pan, Zhiqi Ge, Minghe Gao, Wei Ji, Wenqiao Zhang, Tat-Seng Chua, Siliang Tang, Hanwang Zhang, and Yueting Zhuang. 2024{\natexlab{c}}.
\newblock \href {https://openreview.net/forum?id=BXY6fe7q31} {Fine-tuning multimodal {LLM}s to follow zero-shot demonstrative instructions}.
\newblock In \emph{The Twelfth International Conference on Learning Representations}.

\bibitem[{Li et~al.(2023)Li, Du, Zhou, Wang, Zhao, and Wen}]{li-etal-2023-pope}
Yifan Li, Yifan Du, Kun Zhou, Jinpeng Wang, Xin Zhao, and Ji-Rong Wen. 2023.
\newblock \href {https://doi.org/10.18653/v1/2023.emnlp-main.20} {Evaluating object hallucination in large vision-language models}.
\newblock In \emph{Proceedings of the 2023 Conference on Empirical Methods in Natural Language Processing}, pages 292--305. Association for Computational Linguistics.

\bibitem[{Li and Zhang(2023{\natexlab{a}})}]{li-zhang-2023-mcc}
Zhi Li and Yin Zhang. 2023{\natexlab{a}}.
\newblock \href {https://doi.org/10.18653/v1/2023.emnlp-main.18} {Cultural concept adaptation on multimodal reasoning}.
\newblock In \emph{Proceedings of the 2023 Conference on Empirical Methods in Natural Language Processing}, pages 262--276. Association for Computational Linguistics.

\bibitem[{Li and Zhang(2023{\natexlab{b}})}]{li-zhang-2023-cultural}
Zhi Li and Yin Zhang. 2023{\natexlab{b}}.
\newblock \href {https://doi.org/10.18653/v1/2023.emnlp-main.18} {Cultural concept adaptation on multimodal reasoning}.
\newblock In \emph{Proceedings of the 2023 Conference on Empirical Methods in Natural Language Processing}, pages 262--276. Association for Computational Linguistics.

\bibitem[{Lin et~al.(2024)Lin, Yin, Ping, Molchanov, Shoeybi, and Han}]{lin2024vila}
Ji~Lin, Hongxu Yin, Wei Ping, Pavlo Molchanov, Mohammad Shoeybi, and Song Han. 2024.
\newblock \href {https://openaccess.thecvf.com/content/CVPR2024/html/Lin_VILA_On_Pre-training_for_Visual_Language_Models_CVPR_2024_paper.html} {Vila: On pre-training for visual language models}.
\newblock In \emph{Proceedings of the IEEE/CVF Conference on Computer Vision and Pattern Recognition}, pages 26689--26699.

\bibitem[{Liu et~al.(2021{\natexlab{a}})Liu, Bugliarello, Ponti, Reddy, Collier, and Elliott}]{liu2021visually}
Fangyu Liu, Emanuele Bugliarello, Edoardo~Maria Ponti, Siva Reddy, Nigel Collier, and Desmond Elliott. 2021{\natexlab{a}}.
\newblock \href {https://aclanthology.org/2021.emnlp-main.818/} {Visually grounded reasoning across languages and cultures}.
\newblock In \emph{Proceedings of the 2021 Conference on Empirical Methods in Natural Language Processing}, pages 10467--10485.

\bibitem[{Liu et~al.(2021{\natexlab{b}})Liu, Bugliarello, Ponti, Reddy, Collier, and Elliott}]{liu-etal-2021-visually}
Fangyu Liu, Emanuele Bugliarello, Edoardo~Maria Ponti, Siva Reddy, Nigel Collier, and Desmond Elliott. 2021{\natexlab{b}}.
\newblock \href {https://doi.org/10.18653/v1/2021.emnlp-main.818} {Visually grounded reasoning across languages and cultures}.
\newblock In \emph{Proceedings of the 2021 Conference on Empirical Methods in Natural Language Processing}, pages 10467--10485, Online and Punta Cana, Dominican Republic. Association for Computational Linguistics.

\bibitem[{Liu et~al.(2024{\natexlab{a}})Liu, Li, Li, and Lee}]{liu2024improved}
Haotian Liu, Chunyuan Li, Yuheng Li, and Yong~Jae Lee. 2024{\natexlab{a}}.
\newblock \href {https://openaccess.thecvf.com/content/CVPR2024/html/Liu_Improved_Baselines_with_Visual_Instruction_Tuning_CVPR_2024_paper.html} {Improved baselines with visual instruction tuning}.
\newblock In \emph{Proceedings of the IEEE/CVF Conference on Computer Vision and Pattern Recognition}, pages 26296--26306.

\bibitem[{Liu et~al.(2024{\natexlab{b}})Liu, Li, Li, Li, Zhang, Shen, and Lee}]{liu2024llavanext}
Haotian Liu, Chunyuan Li, Yuheng Li, Bo~Li, Yuanhan Zhang, Sheng Shen, and Yong~Jae Lee. 2024{\natexlab{b}}.
\newblock \href {https://llava-vl.github.io/blog/2024-01-30-llava-next/} {Llava-next: Improved reasoning, ocr, and world knowledge}.

\bibitem[{Liu et~al.(2024{\natexlab{c}})Liu, Chu, Zang, Wei, Dong, Zhang, Liang, Xiong, Qiao, Lin et~al.}]{liu2024mmdu}
Ziyu Liu, Tao Chu, Yuhang Zang, Xilin Wei, Xiaoyi Dong, Pan Zhang, Zijian Liang, Yuanjun Xiong, Yu~Qiao, Dahua Lin, et~al. 2024{\natexlab{c}}.
\newblock \href {https://arxiv.org/abs/2406.11833v1} {Mmdu: A multi-turn multi-image dialog understanding benchmark and instruction-tuning dataset for lvlms}.
\newblock \emph{arXiv preprint arXiv:2406.11833}.

\bibitem[{Majid et~al.(2015)Majid, Jordan, and Dunn}]{majid2015semantic}
Asifa Majid, Fiona Jordan, and Michael Dunn. 2015.
\newblock Semantic systems in closely related languages.

\bibitem[{McKinzie et~al.(2024)McKinzie, Gan, Fauconnier, Dodge, Zhang, Dufter, Shah, Du, Peng, Weers et~al.}]{mckinzie2024mm1}
Brandon McKinzie, Zhe Gan, Jean-Philippe Fauconnier, Sam Dodge, Bowen Zhang, Philipp Dufter, Dhruti Shah, Xianzhi Du, Futang Peng, Floris Weers, et~al. 2024.
\newblock \href {https://arxiv.org/abs/2403.09611} {Mm1: Methods, analysis \& insights from multimodal llm pre-training}.
\newblock \emph{arXiv preprint arXiv:2403.09611}.

\bibitem[{{Meta}(2024)}]{llama3.2}
{Meta}. 2024.
\newblock \href {https://ai.meta.com/blog/llama-3-2-connect-2024-vision-edge-mobile-devices/} {Llama 3.2: Revolutionizing edge ai and vision with open, customizable models}.

\bibitem[{Miquel-Ribé and Laniado(2018)}]{10.3389/fphy.2018.00054}
Marc Miquel-Ribé and David Laniado. 2018.
\newblock \href {https://doi.org/10.3389/fphy.2018.00054} {Wikipedia culture gap: Quantifying content imbalances across 40 language editions}.
\newblock \emph{Frontiers in Physics}, 6.

\bibitem[{Nayak et~al.(2024)Nayak, Jain, Awal, Reddy, van Steenkiste, Hendricks, Sta{\'n}czak, and Agrawal}]{nayak2024benchmarking}
Shravan Nayak, Kanishk Jain, Rabiul Awal, Siva Reddy, Sjoerd van Steenkiste, Lisa~Anne Hendricks, Karolina Sta{\'n}czak, and Aishwarya Agrawal. 2024.
\newblock \href {https://arxiv.org/abs/2407.10920} {Benchmarking vision language models for cultural understanding}.
\newblock \emph{arXiv preprint arXiv:2407.10920}.

\bibitem[{Neeman et~al.(2023)Neeman, Aharoni, Honovich, Choshen, Szpektor, and Abend}]{neeman2023disentqa}
Ella Neeman, Roee Aharoni, Or~Honovich, Leshem Choshen, Idan Szpektor, and Omri Abend. 2023.
\newblock \href {https://doi.org/10.18653/v1/2023.acl-long.559} {{D}isent{QA}: Disentangling parametric and contextual knowledge with counterfactual question answering}.
\newblock In \emph{Proceedings of the 61st Annual Meeting of the Association for Computational Linguistics (Volume 1: Long Papers)}, pages 10056--10070. Association for Computational Linguistics.

\bibitem[{Nwatu et~al.(2023)Nwatu, Ignat, and Mihalcea}]{nwatu-etal-2023-bridging}
Joan Nwatu, Oana Ignat, and Rada Mihalcea. 2023.
\newblock \href {https://doi.org/10.18653/v1/2023.emnlp-main.660} {Bridging the digital divide: Performance variation across socio-economic factors in vision-language models}.
\newblock In \emph{Proceedings of the 2023 Conference on Empirical Methods in Natural Language Processing}, pages 10686--10702. Association for Computational Linguistics.

\bibitem[{Pouget et~al.(2024)Pouget, Beyer, Bugliarello, Wang, Steiner, Zhai, and Alabdulmohsin}]{pouget2024no-filter}
Ang{\'e}line Pouget, Lucas Beyer, Emanuele Bugliarello, Xiao Wang, Andreas~Peter Steiner, Xiaohua Zhai, and Ibrahim Alabdulmohsin. 2024.
\newblock \href {https://arxiv.org/abs/2405.13777} {No filter: Cultural and socioeconomic diversity in contrastive vision-language models}.
\newblock \emph{arXiv preprint arXiv:2405.13777}.

\bibitem[{Radford et~al.(2021)Radford, Kim, Hallacy, Ramesh, Goh, Agarwal, Sastry, Askell, Mishkin, Clark et~al.}]{radford2021learning}
Alec Radford, Jong~Wook Kim, Chris Hallacy, Aditya Ramesh, Gabriel Goh, Sandhini Agarwal, Girish Sastry, Amanda Askell, Pamela Mishkin, Jack Clark, et~al. 2021.
\newblock \href {https://proceedings.mlr.press/v139/radford21a} {Learning transferable visual models from natural language supervision}.
\newblock In \emph{International conference on machine learning}, pages 8748--8763. PMLR.

\bibitem[{Raudenbush(1994)}]{raudenbush1994random}
Stephen~W Raudenbush. 1994.
\newblock \href {https://psycnet.apa.org/record/1993-99100-019} {Random effects models}.
\newblock \emph{The handbook of research synthesis}, 421(3.6).

\bibitem[{Richards et~al.(2024)Richards, Kirichenko, Bouchacourt, and Ibrahim}]{richards2024does}
Megan Richards, Polina Kirichenko, Diane Bouchacourt, and Mark Ibrahim. 2024.
\newblock \href {https://openreview.net/forum?id=rhaQbS3K3R} {Does progress on object recognition benchmarks improve generalization on crowdsourced, global data?}
\newblock In \emph{The Twelfth International Conference on Learning Representations}.

\bibitem[{Romero et~al.(2024)Romero, Lyu, Wibowo, Lynn, Hamed, Kishore, Mandal, Dragonetti, Abzaliev, Tonja et~al.}]{romero2024cvqa}
David Romero, Chenyang Lyu, Haryo~Akbarianto Wibowo, Teresa Lynn, Injy Hamed, Aditya~Nanda Kishore, Aishik Mandal, Alina Dragonetti, Artem Abzaliev, Atnafu~Lambebo Tonja, et~al. 2024.
\newblock \href {https://arxiv.org/abs/2406.05967} {Cvqa: Culturally-diverse multilingual visual question answering benchmark}.
\newblock \emph{arXiv preprint arXiv:2406.05967}.

\bibitem[{Salinas and Morstatter(2024)}]{salinas-morstatter-2024-butterfly}
Abel Salinas and Fred Morstatter. 2024.
\newblock \href {https://doi.org/10.18653/v1/2024.findings-acl.275} {The butterfly effect of altering prompts: How small changes and jailbreaks affect large language model performance}.
\newblock In \emph{Findings of the Association for Computational Linguistics ACL 2024}, pages 4629--4651. Association for Computational Linguistics.

\bibitem[{Schielzeth et~al.(2020)Schielzeth, Dingemanse, Nakagawa, Westneat, Allegue, Teplitsky, Réale, Dochtermann, Garamszegi, and Araya-Ajoy}]{Schielzeth_mixed_effects}
Holger Schielzeth, Niels~J. Dingemanse, Shinichi Nakagawa, David~F. Westneat, Hassen Allegue, Céline Teplitsky, Denis Réale, Ned~A. Dochtermann, László~Zsolt Garamszegi, and Yimen~G. Araya-Ajoy. 2020.
\newblock \href {https://doi.org/https://doi.org/10.1111/2041-210X.13434} {Robustness of linear mixed-effects models to violations of distributional assumptions}.
\newblock \emph{Methods in Ecology and Evolution}, 11(9):1141--1152.

\bibitem[{Shankar et~al.(2017)Shankar, Halpern, Breck, Atwood, Wilson, and Sculley}]{46553}
Shreya Shankar, Yoni Halpern, Eric Breck, James Atwood, Jimbo Wilson, and D.~Sculley. 2017.
\newblock \href {https://arxiv.org/abs/1711.08536} {No classification without representation: Assessing geodiversity issues in open data sets for the developing world}.
\newblock In \emph{NIPS 2017 workshop: Machine Learning for the Developing World}.

\bibitem[{Sharma et~al.(2024)Sharma, Saxon, and Wang}]{sharma2024losing}
Aditya Sharma, Michael Saxon, and William~Yang Wang. 2024.
\newblock \href {https://arxiv.org/abs/2406.16851} {Losing visual needles in image haystacks: Vision language models are easily distracted in short and long contexts}.
\newblock \emph{arXiv preprint arXiv:2406.16851}.

\bibitem[{Shekhar et~al.(2017)Shekhar, Pezzelle, Klimovich, Herbelot, Nabi, Sangineto, and Bernardi}]{shekhar2017foil_acl}
Ravi Shekhar, Sandro Pezzelle, Yauhen Klimovich, Aurelie Herbelot, Moin Nabi, Enver Sangineto, and Raffaella Bernardi. 2017.
\newblock \href {https://aclanthology.org/P17-1024/} {"foil it! find one mismatch between image and language caption"}.
\newblock In \emph{Proceedings of the 55th Annual Meeting of the Association for Computational Linguistics (ACL) (Volume 1: Long Papers)}, pages 255--265.

\bibitem[{Soldaini et~al.(2024)Soldaini, Kinney, Bhagia, Schwenk, Atkinson, Authur, Bogin, Chandu, Dumas, Elazar, Hofmann, Jha, Kumar, Lucy, Lyu, Lambert, Magnusson, Morrison, Muennighoff, Naik, Nam, Peters, Ravichander, Richardson, Shen, Strubell, Subramani, Tafjord, Walsh, Zettlemoyer, Smith, Hajishirzi, Beltagy, Groeneveld, Dodge, and Lo}]{soldaini-etal-2024-dolma}
Luca Soldaini, Rodney Kinney, Akshita Bhagia, Dustin Schwenk, David Atkinson, Russell Authur, Ben Bogin, Khyathi Chandu, Jennifer Dumas, Yanai Elazar, Valentin Hofmann, Ananya Jha, Sachin Kumar, Li~Lucy, Xinxi Lyu, Nathan Lambert, Ian Magnusson, Jacob Morrison, Niklas Muennighoff, Aakanksha Naik, Crystal Nam, Matthew Peters, Abhilasha Ravichander, Kyle Richardson, Zejiang Shen, Emma Strubell, Nishant Subramani, Oyvind Tafjord, Evan Walsh, Luke Zettlemoyer, Noah Smith, Hannaneh Hajishirzi, Iz~Beltagy, Dirk Groeneveld, Jesse Dodge, and Kyle Lo. 2024.
\newblock \href {https://doi.org/10.18653/v1/2024.acl-long.840} {Dolma: an open corpus of three trillion tokens for language model pretraining research}.
\newblock In \emph{Proceedings of the 62nd Annual Meeting of the Association for Computational Linguistics (Volume 1: Long Papers)}, pages 15725--15788. Association for Computational Linguistics.

\bibitem[{Song et~al.(2024)Song, Chen, Chen, Yu, Wan, and Wang}]{song2024milebench}
Dingjie Song, Shunian Chen, Guiming~Hardy Chen, Fei Yu, Xiang Wan, and Benyou Wang. 2024.
\newblock \href {https://arxiv.org/abs/2404.18532} {Milebench: Benchmarking mllms in long context}.
\newblock \emph{arXiv preprint arXiv:2404.18532}.

\bibitem[{Tao et~al.(2023)Tao, Viberg, Baker, and Kizilcec}]{tao2023auditing}
Yan Tao, Olga Viberg, Ryan~S Baker, and Rene~F Kizilcec. 2023.
\newblock \href {https://arxiv.org/abs/2311.14096} {Auditing and mitigating cultural bias in llms}.
\newblock \emph{arXiv preprint arXiv:2311.14096}.

\bibitem[{Team et~al.(2023)Team, Anil, Borgeaud, Wu, Alayrac, Yu, Soricut, Schalkwyk, Dai, Hauth et~al.}]{team2023gemini}
Gemini Team, Rohan Anil, Sebastian Borgeaud, Yonghui Wu, Jean-Baptiste Alayrac, Jiahui Yu, Radu Soricut, Johan Schalkwyk, Andrew~M Dai, Anja Hauth, et~al. 2023.
\newblock \href {https://arxiv.org/abs/2312.11805} {Gemini: a family of highly capable multimodal models}.
\newblock \emph{arXiv preprint arXiv:2312.11805}.

\bibitem[{Thapliyal et~al.(2022)Thapliyal, Pont~Tuset, Chen, and Soricut}]{thapliyal-etal-2022-crossmodal3600}
Ashish~V. Thapliyal, Jordi Pont~Tuset, Xi~Chen, and Radu Soricut. 2022.
\newblock \href {https://doi.org/10.18653/v1/2022.emnlp-main.45} {Crossmodal-3600: A massively multilingual multimodal evaluation dataset}.
\newblock In \emph{Proceedings of the 2022 Conference on Empirical Methods in Natural Language Processing}, pages 715--729. Association for Computational Linguistics.

\bibitem[{Thissen et~al.(2002)Thissen, Steinberg, and Kuang}]{thissen2002quick}
David Thissen, Lynne Steinberg, and Daniel Kuang. 2002.
\newblock \href {https://journals.sagepub.com/doi/10.3102/10769986027001077} {Quick and easy implementation of the benjamini-hochberg procedure for controlling the false positive rate in multiple comparisons}.
\newblock \emph{Journal of educational and behavioral statistics}, 27(1):77--83.

\bibitem[{Tsimpoukelli et~al.(2021)Tsimpoukelli, Menick, Cabi, Eslami, Vinyals, and Hill}]{tsimpoukelli2021multimodal}
Maria Tsimpoukelli, Jacob~L Menick, Serkan Cabi, SM~Eslami, Oriol Vinyals, and Felix Hill. 2021.
\newblock \href {https://proceedings.neurips.cc/paper/2021/hash/01b7575c38dac42f3cfb7d500438b875-Abstract.html} {Multimodal few-shot learning with frozen language models}.
\newblock \emph{Advances in Neural Information Processing Systems}, 34:200--212.

\bibitem[{Urailertprasert et~al.(2024)Urailertprasert, Limkonchotiwat, Suwajanakorn, and Nutanong}]{urailertprasert-etal-2024-seavqa}
Norawit Urailertprasert, Peerat Limkonchotiwat, Supasorn Suwajanakorn, and Sarana Nutanong. 2024.
\newblock \href {https://doi.org/10.18653/v1/2024.alvr-1.15} {{SEA}-{VQA}: {S}outheast {A}sian cultural context dataset for visual question answering}.
\newblock In \emph{Proceedings of the 3rd Workshop on Advances in Language and Vision Research (ALVR)}, pages 173--185. Association for Computational Linguistics.

\bibitem[{Wang et~al.(2024{\natexlab{a}})Wang, Bai, Tan, Wang, Fan, Bai, Chen, Liu, Wang, Ge, Fan, Dang, Du, Ren, Men, Liu, Zhou, Zhou, and Lin}]{Qwen2VL}
Peng Wang, Shuai Bai, Sinan Tan, Shijie Wang, Zhihao Fan, Jinze Bai, Keqin Chen, Xuejing Liu, Jialin Wang, Wenbin Ge, Yang Fan, Kai Dang, Mengfei Du, Xuancheng Ren, Rui Men, Dayiheng Liu, Chang Zhou, Jingren Zhou, and Junyang Lin. 2024{\natexlab{a}}.
\newblock \href {https://arxiv.org/abs/2409.12191} {Qwen2-vl: Enhancing vision-language model's perception of the world at any resolution}.
\newblock \emph{arXiv preprint arXiv:2409.12191}.

\bibitem[{Wang et~al.(2024{\natexlab{b}})Wang, Bai, Tan, Wang, Fan, Bai, Chen, Liu, Wang, Ge et~al.}]{wang2024qwen2}
Peng Wang, Shuai Bai, Sinan Tan, Shijie Wang, Zhihao Fan, Jinze Bai, Keqin Chen, Xuejing Liu, Jialin Wang, Wenbin Ge, et~al. 2024{\natexlab{b}}.
\newblock \href {https://arxiv.org/abs/2409.12191} {Qwen2-vl: Enhancing vision-language model's perception of the world at any resolution}.
\newblock \emph{arXiv preprint arXiv:2409.12191}.

\bibitem[{Wang et~al.(2024{\natexlab{c}})Wang, Zhang, Ren, Duan, Li, Liu, Hu, Chen, Zhang, Lu et~al.}]{wang2024needle}
Weiyun Wang, Shuibo Zhang, Yiming Ren, Yuchen Duan, Tiantong Li, Shuo Liu, Mengkang Hu, Zhe Chen, Kaipeng Zhang, Lewei Lu, et~al. 2024{\natexlab{c}}.
\newblock \href {https://arxiv.org/abs/2406.07230} {Needle in a multimodal haystack}.
\newblock \emph{arXiv preprint arXiv:2406.07230}.

\bibitem[{Wei et~al.(2023)Wei, Chen, Chen, Hu, Zhang, Fu, Ritter, and Chen}]{wei2023uniir}
Cong Wei, Yang Chen, Haonan Chen, Hexiang Hu, Ge~Zhang, Jie Fu, Alan Ritter, and Wenhu Chen. 2023.
\newblock Uniir: Training and benchmarking universal multimodal information retrievers.
\newblock \emph{arXiv preprint arXiv:2311.17136}.

\bibitem[{Wu et~al.(2024)Wu, Zhang, Zhang, Chen, Liao, Wang, Li, Sun, Yan, Zhai, and Lin}]{wu2024qbench}
Haoning Wu, Zicheng Zhang, Erli Zhang, Chaofeng Chen, Liang Liao, Annan Wang, Chunyi Li, Wenxiu Sun, Qiong Yan, Guangtao Zhai, and Weisi Lin. 2024.
\newblock \href {https://openreview.net/forum?id=0V5TVt9bk0} {Q-bench: A benchmark for general-purpose foundation models on low-level vision}.
\newblock In \emph{The Twelfth International Conference on Learning Representations}.

\bibitem[{Xue et~al.(2024)Xue, Shu, Awadalla, Wang, Yan, Purushwalkam, Zhou, Prabhu, Dai, Ryoo et~al.}]{xue2024xgen}
Le~Xue, Manli Shu, Anas Awadalla, Jun Wang, An~Yan, Senthil Purushwalkam, Honglu Zhou, Viraj Prabhu, Yutong Dai, Michael~S Ryoo, et~al. 2024.
\newblock \href {https://arxiv.org/abs/2408.08872} {xgen-mm (blip-3): A family of open large multimodal models}.
\newblock \emph{arXiv preprint arXiv:2408.08872}.

\bibitem[{Ye et~al.(2024)Ye, Xu, Liu, Hu, Yan, Qian, Zhang, Huang, and Zhou}]{ye2024mplug}
Jiabo Ye, Haiyang Xu, Haowei Liu, Anwen Hu, Ming Yan, Qi~Qian, Ji~Zhang, Fei Huang, and Jingren Zhou. 2024.
\newblock \href {https://arxiv.org/abs/2408.04840} {mplug-owl3: Towards long image-sequence understanding in multi-modal large language models}.
\newblock \emph{arXiv preprint arXiv:2408.04840}.

\bibitem[{Yin et~al.(2023)Yin, Gao, Thattai, Johnston, and Chang}]{yin2023givl}
Da~Yin, Feng Gao, Govind Thattai, Michael Johnston, and Kai-Wei Chang. 2023.
\newblock \href {https://openaccess.thecvf.com/content/CVPR2023/html/Yin_GIVL_Improving_Geographical_Inclusivity_of_Vision-Language_Models_With_Pre-Training_Methods_CVPR_2023_paper.html} {Givl: Improving geographical inclusivity of vision-language models with pre-training methods}.
\newblock In \emph{Proceedings of the IEEE/CVF Conference on Computer Vision and Pattern Recognition}, pages 10951--10961.

\bibitem[{Yin et~al.(2021)Yin, Li, Hu, Peng, and Chang}]{yin2021gd-vcr}
Da~Yin, Liunian~Harold Li, Ziniu Hu, Nanyun Peng, and Kai-Wei Chang. 2021.
\newblock \href {https://aclanthology.org/2021.emnlp-main.162} {{Broaden the Vision: Geo-Diverse Visual Commonsense Reasoning}}.
\newblock In \emph{EMNLP}.

\bibitem[{Yu et~al.(2024)Yu, Yang, Li, Wang, Lin, Liu, Wang, and Wang}]{yu2024mm}
Weihao Yu, Zhengyuan Yang, Linjie Li, Jianfeng Wang, Kevin Lin, Zicheng Liu, Xinchao Wang, and Lijuan Wang. 2024.
\newblock Mm-vet: Evaluating large multimodal models for integrated capabilities.
\newblock In \emph{International conference on machine learning}. PMLR.

\bibitem[{Zhai et~al.(2023)Zhai, Mustafa, Kolesnikov, and Beyer}]{zhai2023sigmoid}
Xiaohua Zhai, Basil Mustafa, Alexander Kolesnikov, and Lucas Beyer. 2023.
\newblock \href {https://openaccess.thecvf.com/content/ICCV2023/html/Zhai_Sigmoid_Loss_for_Language_Image_Pre-Training_ICCV_2023_paper.html} {Sigmoid loss for language image pre-training}.
\newblock In \emph{Proceedings of the IEEE/CVF International Conference on Computer Vision}, pages 11975--11986.

\bibitem[{Zhang et~al.(2024)Zhang, Dong, Zang, Cao, Qian, Chen, Guo, Duan, Wang, Ouyang et~al.}]{zhang2024internlm}
Pan Zhang, Xiaoyi Dong, Yuhang Zang, Yuhang Cao, Rui Qian, Lin Chen, Qipeng Guo, Haodong Duan, Bin Wang, Linke Ouyang, et~al. 2024.
\newblock \href {https://arxiv.org/abs/2407.03320} {Internlm-xcomposer-2.5: A versatile large vision language model supporting long-contextual input and output}.
\newblock \emph{arXiv preprint arXiv:2407.03320}.

\bibitem[{Zhang et~al.(2023)Zhang, Li, Hauer, Shi, and Kondrak}]{zhang2023dont}
Xiang Zhang, Senyu Li, Bradley Hauer, Ning Shi, and Grzegorz Kondrak. 2023.
\newblock \href {https://openreview.net/forum?id=WiKLXsWzBy} {Don{\textquoteright}t trust chat{GPT} when your question is not in english: A study of multilingual abilities and types of {LLM}s}.
\newblock In \emph{The 2023 Conference on Empirical Methods in Natural Language Processing}.

\bibitem[{Zong et~al.(2024)Zong, Bohdal, and Hospedales}]{zong2024vl}
Yongshuo Zong, Ondrej Bohdal, and Timothy Hospedales. 2024.
\newblock \href {https://arxiv.org/abs/2403.13164} {Vl-icl bench: The devil in the details of benchmarking multimodal in-context learning}.
\newblock \emph{arXiv preprint arXiv:2403.13164}.

\end{thebibliography}
\bibliographystyle{acl_natbib}

\appendix

\section{Dataset Collection}\label{sec:data-collection}
\begin{figure}[ht!]
    \centering
    \includegraphics[width=\linewidth]{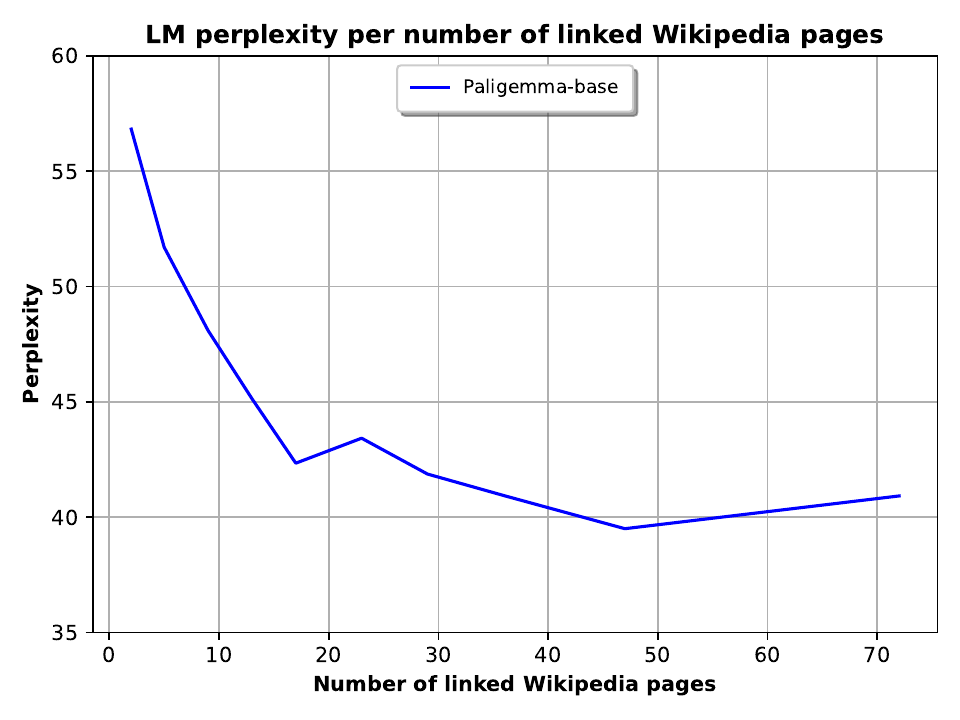}
    \caption{Perplexity of Wikipedia image captions vs the number of linked languages for the page including that image.}
    \label{fig:perplexity}
\end{figure}

\subsection{Model Perplexity vs Number of Linked Languages}
\label{sec:appendix-perplexity}
During our dataset construction, we focused on more culture-specific concepts by filtering out concepts based on the number of languages in Wikipedia. 
Intuitively, if a concept is not present on Wikipedia for a particular language, it is likely that the concept is less common. 
This is particularly impactful in the case of VLMs \citep{laurençon2024idefics2, deitke2024molmo} and their respective backbone LLMs \citep{groeneveld-etal-2024-olmo, soldaini-etal-2024-dolma} where Wikipedia is often considered a high data resource.

To showcase this, we measured the perplexity of Paligemma when completing captions from images of Wikipedia containing concepts with varying number of links.
\cref{fig:perplexity} shows the perplexity aggregated for different number of linked languages in Wikipedia. 
We observe a clear trend, where the more a concept is represented in multiple languages, the lower the perplexity of models.

\subsection{Human Annotation}\label{sec:human_anntotation}
\begin{figure}[tb]
    \centering
    \includegraphics[width=\linewidth]{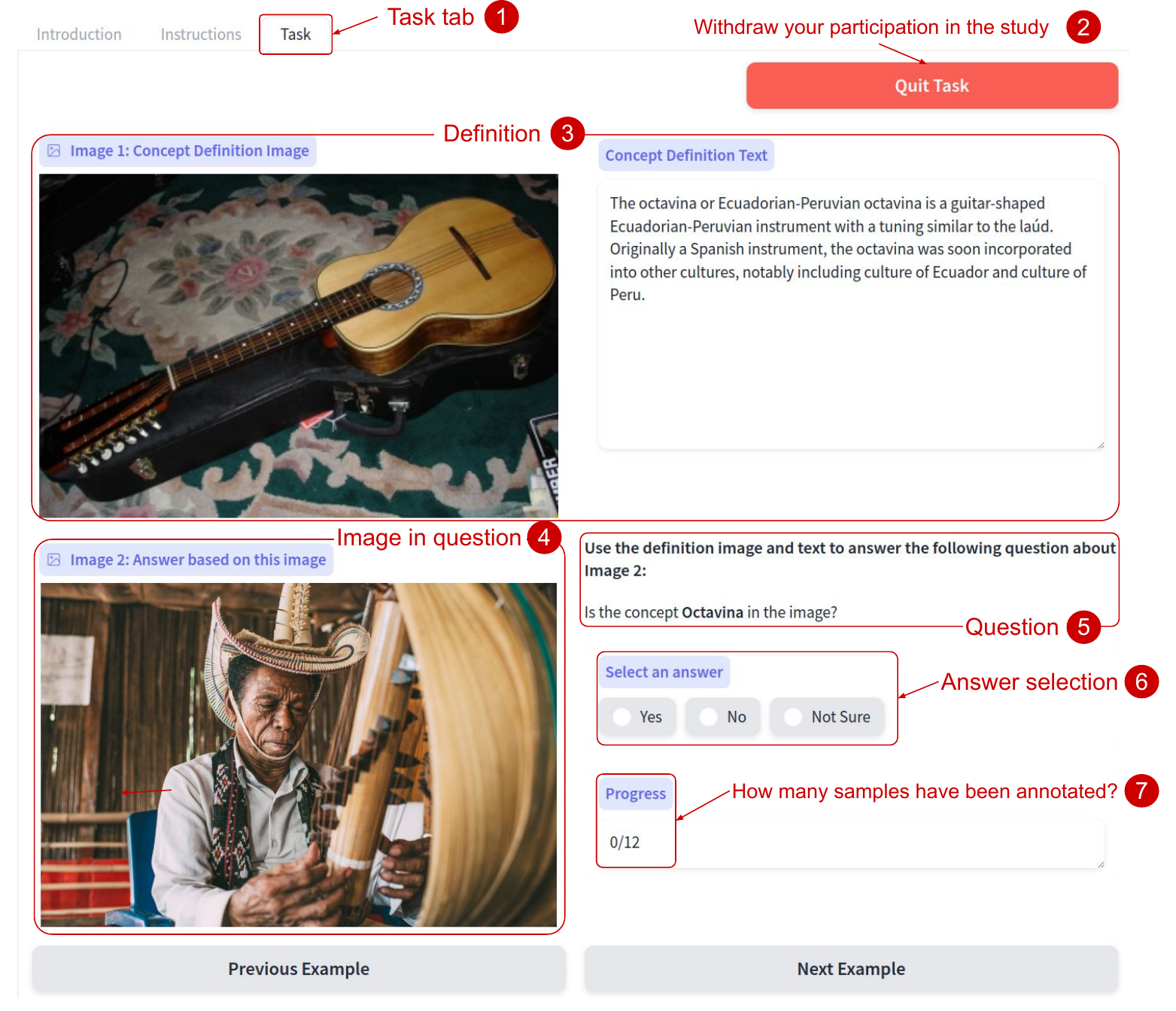}
    \caption{Annotated layout of the annotation interface as shown to the annotators.}
    \label{fig:layout}
\end{figure}
\begin{figure}[tb]
    \centering
    \includegraphics[width=\linewidth]{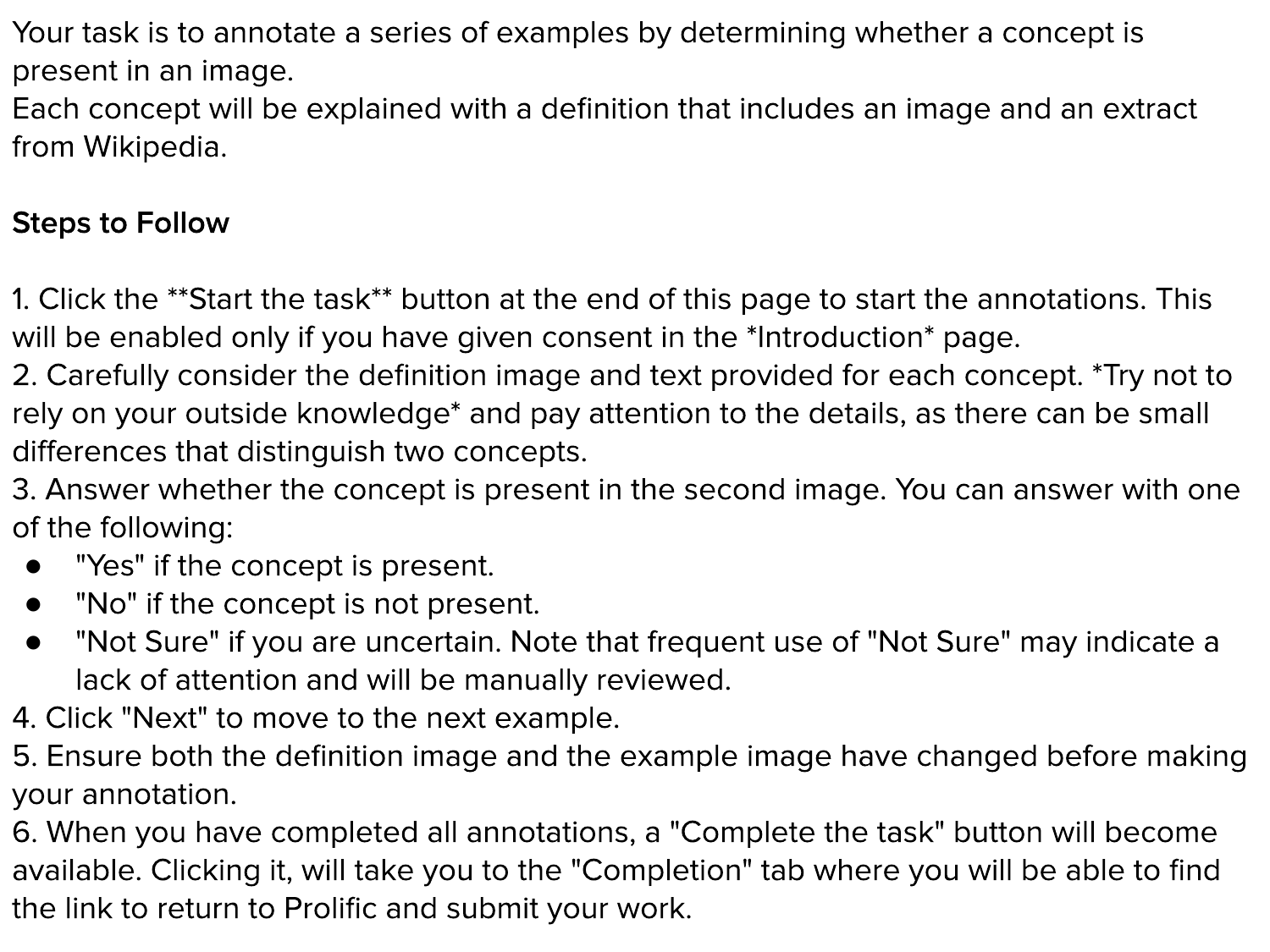}
    \caption{Human annotation instructions.}
    \label{fig:instructions}
\end{figure}
Both the Data Annotation and Human Evaluation were conducted through the Prolific platform.
We developed the annotation interface shown in \cref{fig:layout} using Gradio\footnote{\href{https://www.gradio.app/}{https://www.gradio.app/}}.
For each sample, participants are given an example image and the Wikipedia summary for a concept and are asked to answer if the concept appears in a second image.
For the concepts which are associated with multiple images, we manually select the example image to ensure it is representative of the target concept.
Participants are compensated with 15.00\$/ per hour.

We split our data into batches where each batch consists of examples from the same source language.
We recruited participants through a Qualification Stage with the criteria that their primary language is the source language of the batch and that they are fluent in English.
Participants signed a virtual consent with details about the data collection,
how the data would be used and how they could withdraw if they wished to do so.
During the Qualification stage, we asked participants to annotate 10 samples for which the ground truth was known. 
At the end of the task, we provided participants with corrections for any mistakes and explanations for the correct answer.
To ensure high-quality annotations, we only invite participants who answered at least 70\% of the qualification examples correctly.
Consequently, for the Annotation Stage, we recruit 10 participants per language (with the exception of Tamil for which we recruit 12).
Participants are then asked to annotate the dataset examples. 
Each example is annotated by three participants.
In this task, we also allowed participants to answer with `Not Sure,' which allows us to discard any ambiguous examples (indicated by at least two `Not Sure' or tie across the three possible labels).
Each participant annotated, on average, 64 samples.

The Human Evaluation study (\cref{sec:human_eval}) is set up in a similar way.
For each condition, we recruit 35 new participants who are fluent in English as the only requirement.
Additionally, we ask each participant to annotate 10 samples from different concepts.
Before being shown each sample, participants are also asked to rank on a 5-point Likert scale how familiar they are with the concept that appears in the question.
Finally, we only allow participants to answer with `Yes' and `No'.

\section{Dataset}
\label{sec:appendix-dataset}
\subsection{Dataset Analysis}
\label{sec:appendix-dataset-summary}
\begin{table}[!ht]
    \centering
    \resizebox{0.99\linewidth}{!}{
    \begin{tabular}{lccccc}\toprule
    \bf{Source Language Code} & \bf{ID} & \bf{SW} & \bf{TA} & \bf{TR} & \bf{ZH} \\\midrule
    \# Samples  & 228 & 194 & 230 & 207 & 201 \\
    \# Image Concepts & 34 & 39 & 24 & 27 & 34 \\\bottomrule
    \end{tabular}}
    \caption{Number of samples and image concepts per source language.}
    \label{tab:stats}
\end{table}
\begin{figure}[!ht]
    \centering
    \includegraphics[width=\linewidth]{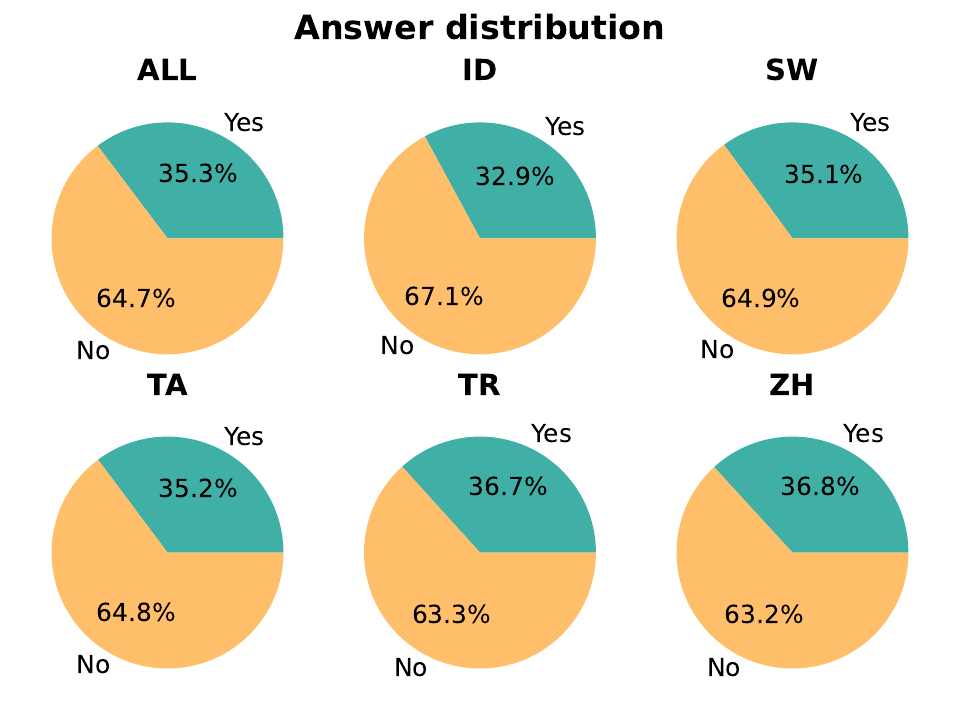}
    \caption{Answer distribution.}
    \label{fig:answer_distribution}
\end{figure}

\begin{figure}[ht!]
    \begin{subfigure}[b]{0.45\textwidth}
        \includegraphics[width=\linewidth]{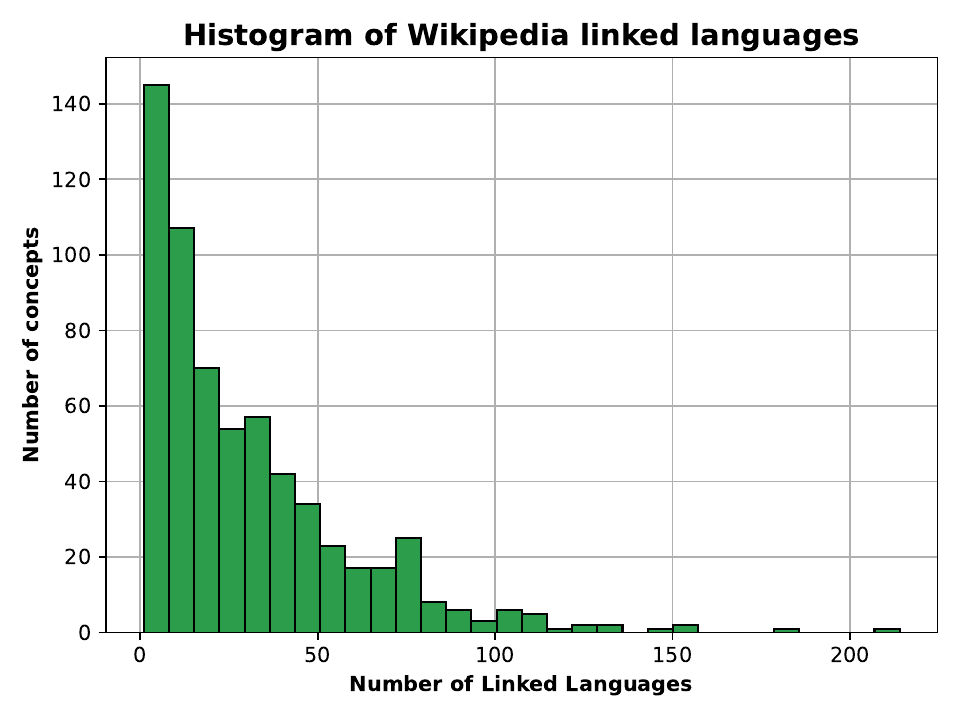}
    \end{subfigure}
    \hfil%
    \begin{subfigure}[b]{0.45\textwidth}
        \includegraphics[width=\linewidth]{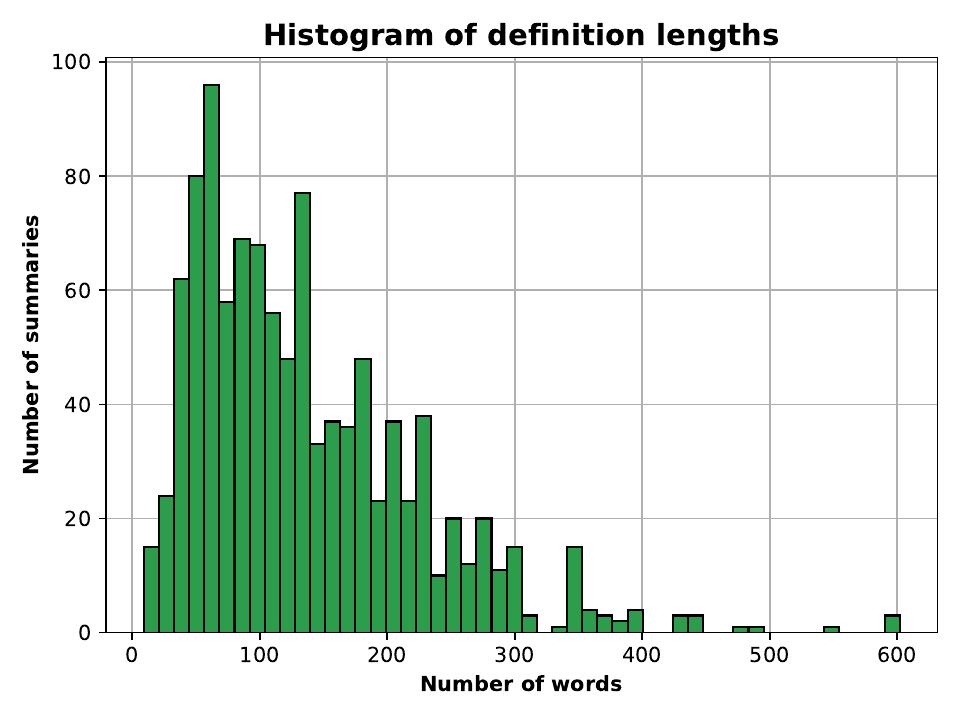}
    \end{subfigure}
    \hfil%
    \begin{subfigure}[b]{0.45\textwidth}
        \includegraphics[width=\linewidth]{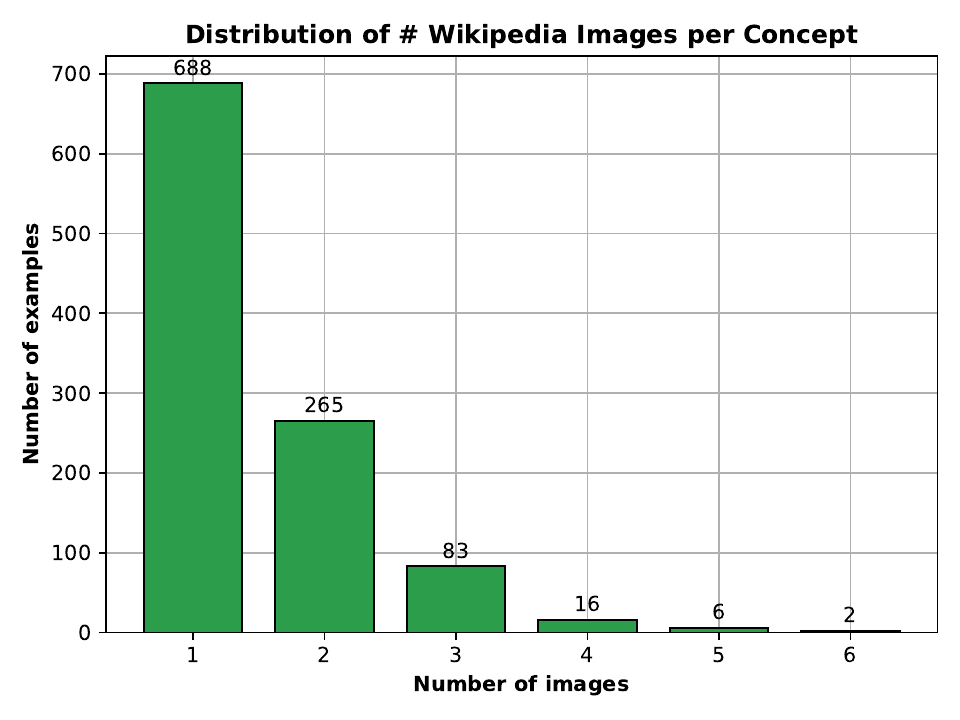}
    \end{subfigure}
    \caption{Statistics for the context of the question concepts. (Top) Histogram of the linked languages in Wikipedia for the concepts in \crope. The majority of the concepts appear in a limited number of languages indicating cultural specificity. (Middle) Histogram of the summary lengths measured as the number of words separated by whitespace. (Bottom) Distribution of \# of Wikipedia images per concept.}
    \label{fig:stats}
\end{figure}
The CROPE dataset consists of 1060 evaluation samples with concepts originating from speakers of five typologically diverse languages \cite{liu-etal-2021-visually}, specifically 228 from Indonesian, 194 from Swahili, 230 from Tamil, 207 from Turkish, and 201 from Chinese.
\cref{fig:answer_distribution} illustrates the answer distribution per source language for each concept.
The dataset is imbalanced, with the majority of the ground truth answers being `No'.
This is done to probe the understanding of the image concept.
The answer distribution is maintained across all source languages in our dataset.

The top part of \cref{fig:stats} illustrates the number of Wikipedia pages, each with a unique language, that is available per question concept (which includes both image and negative concepts). 
Note that the vast majority of the concepts that we have selected appear on Wikipedia in less than 50 languages.
Additionally, the middle part of \cref{fig:stats} shows the distribution of the number of words (separated by whitespace) of the concept summary from Wikipedia.
All summaries can easily fit within the context window of modern VLMs \citep{li2024llava-onevision, laurençon2024idefics2}.
Finally, the bottom part of \cref{fig:stats} depicts the distribution of the number of context images available per concept.

\paragraph{What information does the Wikipedia summary contain that can be used by models to infer the visual appearance of a concept?}
Many Wikipedia summaries for a concept may contain information that is not particularly relevant to its visual appearance.
To quantify the information within the summaries that is useful for identifying a concept, we use GPT-4o \cite{achiam2023gpt4} to extract the text spans from the summary that can be used to infer a plausible visual appearance of a concept. 
More specifically, we give the model demonstrations of extracted text-spans from Wikipedia summaries that provide cues for the concept's visual appearance. 
We then ask the model to extract the spans for the remaining summaries for all concepts in our dataset.
Finally, we compute a word-level overlap between the visually pertinent spans and the entire summary of the concept.

\begin{figure}[tb]
    \centering
    \includegraphics[width=\linewidth]{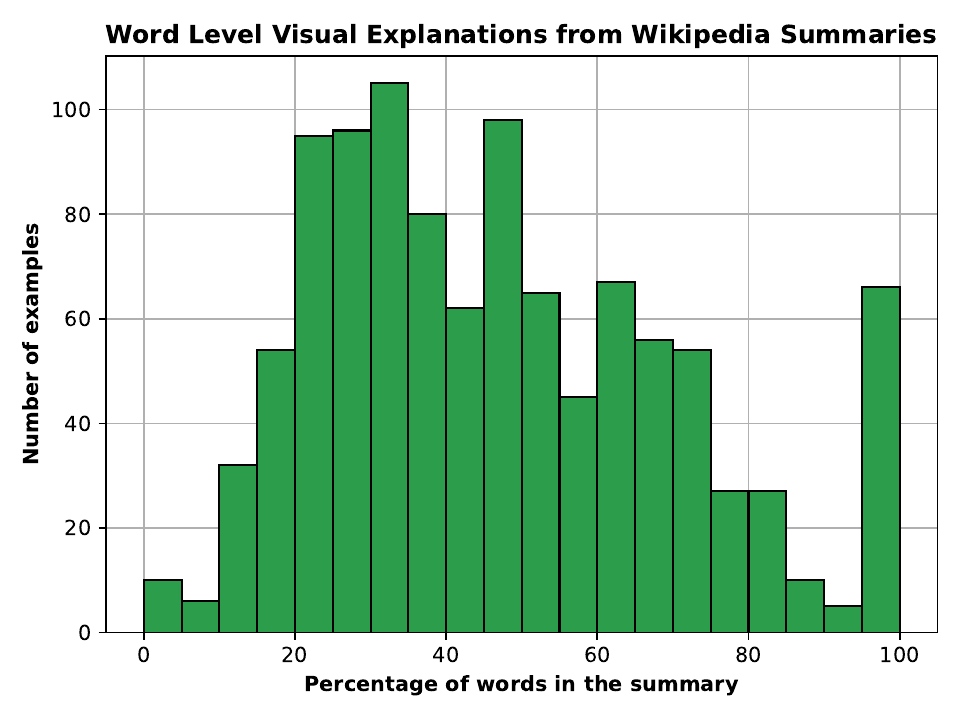}
    \caption{Histogram of word-level overlap percentage between visually relevant text spans and Wikipedia summaries.}
    \label{fig:word-level}
\end{figure}

\cref{fig:word-level} illustrates the histogram of the word-level overlap between the text spans extracted via GPT-4o and the full Wikipedia summary for a concept.
We observe that a significant portion of the summary can be used to infer visual properties of a concept, with even some summaries being identified as fully visually relevant by GPT-4o.
From manual inspection of the extracted spans, these correspond to relatively short summaries that provide the name, the origin, and the functionality of the concept (e.g, \texttt{Chanakhi is a traditional Georgian dish of lamb stew with tomatoes, aubergines, potatoes, greens, and garlic.}).

\subsection{Dataset Examples}\label{sec:appendix-inp-out}

\cref{tab:examples} illustrates examples of input-output pairs, including the text and image context. 
Some of the examples (Example 2 and 4), depict cases where image or the text context is sufficient to answer the question.
Other examples in the table (Example 3 and 5), show cases where the information from a single modality do not provide sufficient cues.

\subsection{License}
We will release the questions and annotations under the CC BY-NC 4.0 license.
The licenses for the different resources used for our dataset are as follows:
Wikipedia texts and images are co-licensed under CC BY-SA 4.0 and the GNU Free Documentation License.
The MaRVL \cite{liu2021visually} text data are released under the CC BY 4.0 license, and all included images are also CC-licensed.
The MCC dataset \cite{li-zhang-2023-cultural} is available on Hugging Face datasets\footnote{\href{https://huggingface.co/datasets}{https://huggingface.co/datasets}} under Apache license 2.0.
All data are provided for non-commercial, research-only purposes.

\section{Implementation Details}
\cref{tab:hf_model_ids} reports the model tags from Hugging Face\footnote{\href{https://huggingface.co/models}{https://huggingface.co/models}} for the open-source/open-weights models.
\cref{tab:prompts} shows the prompts used in this study.
Note that we also follow the guidelines for each model to include system or chat prompts.
Experiments were run on an NVIDIA A40 (40GB), requiring approximately 80 GPU hours to obtain the full results.

When the context includes an example image, and there are multiple available images for the target concept, we use the same image that was selected for the human data collection (\cref{sec:human_anntotation}).
For experiments with a varying number of $k$ of images in context, we always select the first $k$ images and run the experiments for up to three permutations.

\section{Further Results}
\paragraph{Multimodal Performance Analysis}
\begin{table}[tb]
    \centering
    \small
    \begin{tabular}{l|cc}\toprule
    & \multicolumn{2}{c}{\bf{F1 Score}} \\
\bf{Model}    &  \bf{$\notin$ MMDU} & \bf{$\in$ MMDU} \\\midrule
      mPLUG-Owl3   & 68.62 & 68.60 \\
      InternLM-XC-2.5 & 73.52 & 70.02 \\
    XGEN-MM & 80.33 & 84.83 \\
    \bottomrule
    \end{tabular}
    \caption{F1 score under Multimodal context grouped depending on whether the concept is found in the MMDU training data.}
    \label{tab:mmdu}
\end{table}
We notice that three of the models with the strongest relative performance when using multimodal context (XGen-MM, InternLM-XComposer-2.5, mPLUG-Owl3) include the MMDU dataset \cite{liu2024mmdu} in their supervised finetuning data mixture.
MMDU is a recent multi-image instruction following dataset with images from Wikipedia, and dialogs generated by prompting GPT-4o with the relevant images and text information.
We find that only 19.5\% of the concepts in the questions of \crope{} appear in the training set of MMDU. 
\cref{tab:mmdu} shows the F1 score under the multimodal condition aggregated based on whether the image concept appears in MMDU.
We find no evidence of consistent benefit in MMDU concepts, which indicates that the results in the multimodal condition are not due to pure memorization.

\paragraph{Performance of Larger Models}
\cref{tab:larger_model} shows the zero-shot F1 score on the POPE and CROPE test sets as model parameter size increases. We additionally provide results for InternVL-2 \cite{chen2024far}, which is available across multiple sizes.
We observe that while for POPE, performance remains consistent with scale, in CROPE, the F1 score improves. However, only Llama-3.2 (90B) manages to close the gap between culture-specific and common concepts. This is consistent with findings from prior work \cite{pmlr-v202-kandpal23a}, which show that while scaling up improves knowledge of long-tail knowledge, substantial gaps remain, particularly for cases with limited support in pretraining data.

\begin{table}[h!]
    \centering
    \begin{tabular}{lcc}\toprule
    \bf{Model} & \bf{POPE} & \bf{CROPE} \\\midrule
    Llama-3.2-9B & 85.06 & 79.11 \\
    Llama-3.2-90B & 85.60 & 86.32 \\\midrule
    LLaVA-OneVision-7B & 87.66 & 70.68 \\
    LLaVA-OneVision-72B & 85.69 & 77.83 \\\midrule
    MOLMO-7B & 83.88 & 62.50 \\
    MOLMO-72B & 84.70 & 71.41 \\\midrule
    Qwen2-VL-Instruct-7B & 86.91 & 74.06 \\
    Qwen2-VL-Instruct-72B & 86.17 & 77.83 \\
    \bottomrule
    \end{tabular}
    \caption{Zero-shot F1 Score as model size increases.}
    \label{tab:larger_model}
\end{table}

\begin{table*}[tb]
    \centering
    \small
    \begin{tabular}{l|l}
    \toprule
       \bf{ Model} & \bf{Hugging Face Model Name} \\\midrule
        Idefics2 \citep{laurenccon2024matters} &  \texttt{HuggingFaceM4/idefics2-8b} \\
        InternLM-XComposer-2.5 \citep{zhang2024internlm} &  \texttt{internlm/internlm-xcomposer2d5-7b} \\
        Llama-3.2 \citep{llama3.2} &  \texttt{meta-llama/Llama-3.2-9B-Vision-Instruct} \\
        LLaVA-1.5  \citep{liu2024improved} &  \texttt{llava-hf/llava-1.5-7b-hf} \\
        LLaVA-1.5-13B \citep{liu2024improved} & \texttt{llava-hf/llava-1.5-13b-hf} \\
        LLaVA-Next \citep{liu2024llavanext} &  \texttt{llava-hf/llava-v1.6-mistral-7b-hf}\\
        LLaVA-Next-Vicuna-7B \citep{liu2024llavanext} & \texttt{llava-hf/llava-v1.6-vicuna-7b-hf} \\
        LLaVA-Next-Vicuna-13B \citep{liu2024llavanext} & \texttt{llava-hf/llava-v1.6-vicuna-13b-hf} \\
        LLaVA-OneVision \citep{li2024llava-onevision} &  \texttt{lmms-lab/llava-onevision-qwen2-7b-ov}  \\
        Mantis-Idefics2 \citep{jiang2024mantis} &  \texttt{TIGER-Lab/Mantis-8B-Idefics2}\\
        MOLMO \citep{deitke2024molmo} &  \texttt{allenai/Molmo-7B-O-0924} \\
        mPLUG-Owl3 \citep{ye2024mplug} &  \texttt{mPLUG/mPLUG-Owl3-7B-240728} \\
        Paligemma \citep{beyer2024paligemma} &  \texttt{google/paligemma-3b-mix-224} \\
        Phi-3-Vision \citep{abdin2024phi} &  \texttt{microsoft/Phi-3-vision-instruct}\\
        Qwen2-VL \citep{wang2024qwen2} &  \texttt{Qwen/Qwen2-VL-7B-Instruct}\\
        VILA \citep{lin2024vila} &  \texttt{Efficient-Large-Model/Llama-3-VILA1.5-8B}\\
        XGen-MM \citep{xue2024xgen} &  \texttt{Salesforce/xgen-mm-phi3-mini-instruct-interleave-r-v1.5}  \\
        \bottomrule
    \end{tabular}
    \caption{Model details: Hugging Face model names.}
    \label{tab:hf_model_ids}
\end{table*}

\begin{table*}[tb]
\centering
\small
\begin{tabular}{l}
\toprule
 \bf{Prompt Templates}\\ \midrule
{\tt Answer with yes or no: <QUESTION>}\\
{\tt <QUESTION>\textbackslash nAnswer the question using a single word or phrase.}\\
{\tt Look carefully at the previous image and answer the following question with yes or no: <QUESTION>} \\
\bottomrule
\end{tabular}
\caption{Prompt templates. We follow the release guidelines for each model and include the system prompt or chat template as specified. In the Textual context condition, we combine the summary with the question as: {\tt <SUMMARY>\textbackslash n<QUESTION>}.}
\label{tab:prompts}
\end{table*}

\paragraph{Prompt Sensitivity}
\cref{tab:prompt_std} shows the mean and standard deviation of the F1 scores for different prompts under each condition.

\begin{table*}[tb]
    \centering
    \resizebox{\textwidth}{!}{
    \rowcolors{2}{gray!25}{white}
    \begin{tabular}{lcccc}\toprule
\textbf{Model} & \textbf{Zero-Shot} & \textbf{Textual Context} & \textbf{Visual Context} & \textbf{Multimodal Context} \\\midrule
Idefics2-8B (Mistral-7B) & 70.56 $\pm$ 3.98 & 53.72 $\pm$ 1.16 & 44.11 $\pm$ 6.63 & 54.31 $\pm$ 4.82 \\
InternLM-XComposer-2.5 (InternLM2-7B) & 64.73 $\pm$ 2.87 & 57.76 $\pm$ 1.92 & 66.85 $\pm$ 1.47 & 71.94 $\pm$ 0.72 \\
Llama-3.2-Vision-Instruct-11B (Llama-3.1-8B) & 79.11 $\pm$ 2.77 & 53.22 $\pm$ 8.52 & - & - \\
LLaVA-1.5-7B (Vicuna-7B-v1.5) & 62.31 $\pm$ 2.56 & 53.55 $\pm$ 2.96 & - & - \\
LLaVA-1.5-13B (Vicuna-13B-v1.5) & 61.57 $\pm$ 2.23 & 52.88 $\pm$ 2.55 & - & - \\
LLaVA-NeXT-Mistral-7B (Mistral-7B-Instruct-v0.2) & 64.46 $\pm$ 0.79 & 50.95 $\pm$ 2.82 & - & - \\
LLaVA-NeXT-Vicuna-7B (Vicuna-1.5-7B) & 64.16 $\pm$ 1.47 & 58.37 $\pm$ 4.61 & - & - \\
LLaVA-NeXT-Vicuna-13B (Vicuna-1.5-13B) & 63.88 $\pm$ 2.46 & 46.77 $\pm$ 0.39 & - & - \\
LLaVA-OneVision-7B (Qwen2-7B) & 70.68 $\pm$ 2.36 & 56.82 $\pm$ 2.11 & 55.99 $\pm$ 4.26 & 59.84 $\pm$ 4.84 \\
Mantis 8B-Idefics2 (Mistral-7B) & 70.79 $\pm$ 2.31 & 58.20 $\pm$ 1.98 & 67.38 $\pm$ 1.21 & 66.76 $\pm$ 2.42 \\
MOLMO-7B (OLMo-7B-1124) & 62.50 $\pm$ 4.97 & 57.93 $\pm$ 4.59 & - & - \\
mPLUG-Owl3 (Qwen2-7B) & 74.39 $\pm$ 1.38 & 67.10 $\pm$ 2.44 & 49.54 $\pm$ 7.94 & 72.36 $\pm$ 1.00 \\
Paligemma-3B-mix-224 (Gemma-2B) & 68.89 $\pm$ 3.35 & 28.10 $\pm$ 0.45 & - & - \\
Phi-3-Vision-128K-Instruct (Phi3-mini) & 68.94 $\pm$ 2.25 & 62.06 $\pm$ 0.92 & - & - \\
Qwen2-VL-Instruct-7B (Qwen2-7B) & 74.06 $\pm$ 1.58 & 64.73 $\pm$ 2.64 & 77.25 $\pm$ 2.77 & 75.82 $\pm$ 3.58 \\
VILA (Llama-3-8B) & 74.92 $\pm$ 2.32 & 68.73 $\pm$ 1.03 & 70.77 $\pm$ 3.87 & 71.60 $\pm$ 3.15 \\
XGen-MM-Interleaved (Phi3-mini) & 69.24 $\pm$ 0.16 & 52.82 $\pm$ 0.56 & 79.52 $\pm$ 0.68 & 78.75 $\pm$ 0.65 \\
\bottomrule
    \end{tabular}}
    \caption{Mean and standard deviation of F1 score for different prompts.}
    \label{tab:prompt_std}
\end{table*}

\paragraph{Performance per Chapter}
\begin{figure*}
    \centering
    \includegraphics[width=0.8\linewidth]{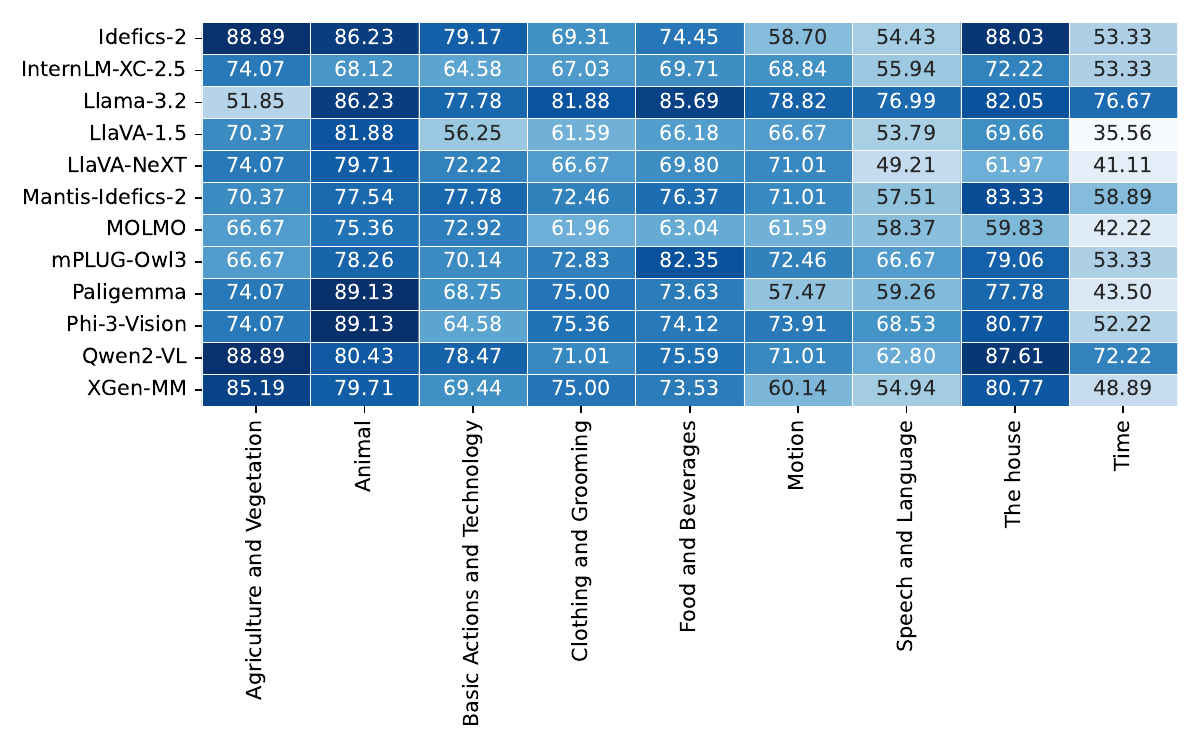}
    \caption{Zero-shot F1 per concept chapter.}
    \label{fig:f1_per_chapter}
\end{figure*}

Each concept is associated with a chapter from the Intercontinental Dictionary Series \cite{borin2013intercontinental}.
\cref{fig:f1_per_chapter} shows the F1 score per concept chapter.
We observe that chapters `Time', that includes concepts about celebrations, and `Speech and Language', which includes concepts about musical instruments and visual art forms, are the most challenging across models. On the other hand, most models score highly on `Agriculture and Vegetation', `Animal', and `The house'.
Overall, we find that different models have different areas of strength and weakness.

\begin{table*}[tb]
     \small
     \centering
\resizebox{\linewidth}{!}{%
     \begin{tabular}{b{0.23\linewidth} b{0.25\linewidth} b{0.23\linewidth} b{0.22\linewidth}}
     \toprule
 \textbf{Exemplar Images} &  \centering \textbf{Wikipedia Summary} &  \centering \textbf{Target Image} &  \textbf{QA and Metadata}
 \\\midrule
\includegraphics[width=0.9\linewidth]{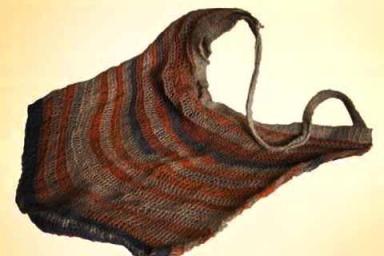} & \small{	
\colorbox{red!30}{Noken} is a traditional Papuan multifunctional knotted or woven bag native to the Western New Guinea region, Indonesia. Its distinctive usage, which involves being hung from the head, is traditionally used to carry various goods, and also children.} &
     \includegraphics[width=0.9\linewidth]{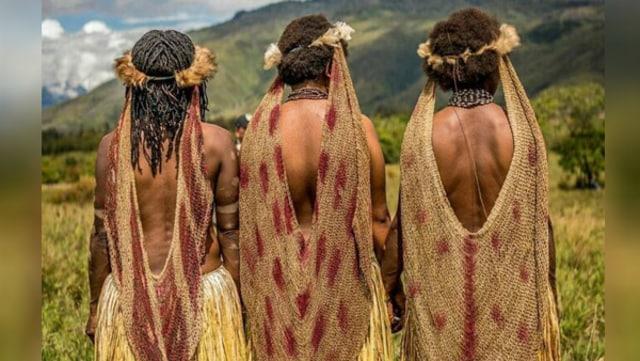} &  
     Q: Is there a noken in the image?
     A: Yes \colorbox{yellow!30}{Noken} \colorbox{mygreen!30}{ID} \colorbox{blue!30}{Clothing and Grooming}
     \\\midrule
\raisebox{0.6\height}{\begin{tabular}{c}
\includegraphics[width=0.9\linewidth]{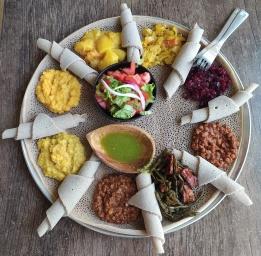}\\
\includegraphics[width=0.9\linewidth]{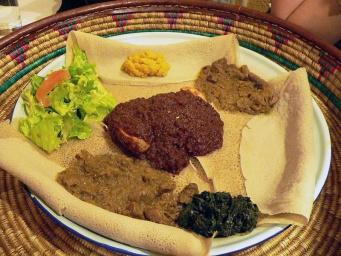}
\end{tabular}} &
\colorbox{red!30}{Injera} is a sour fermented pancake-like flatbread with a slightly spongy texture, traditionally made of teff flour. In Ethiopia and Eritrea, injera is a staple. Injera is central to the dining process in Amhara community, like bread or rice elsewhere and is usually stored in the mesob. & \includegraphics[width=0.9\linewidth]{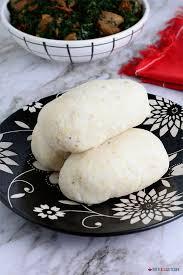} & 
Q: Is there injera in the image? A: No \colorbox{yellow!30}{Ugali} \colorbox{mygreen!30}{SW} \colorbox{blue!30}{Food and Beverages} \\\midrule
\includegraphics[width=0.9\linewidth]{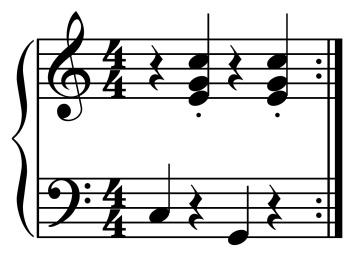} & \small{	
\colorbox{red!30}{Oom-pah}, Oompah or Umpapa is an onomatopoeic term describing the rhythmic sound of a deep brass instrument in combination with the response of other instruments or registers in a band, a form of background ostinato...}
 & \includegraphics[width=0.9\linewidth]{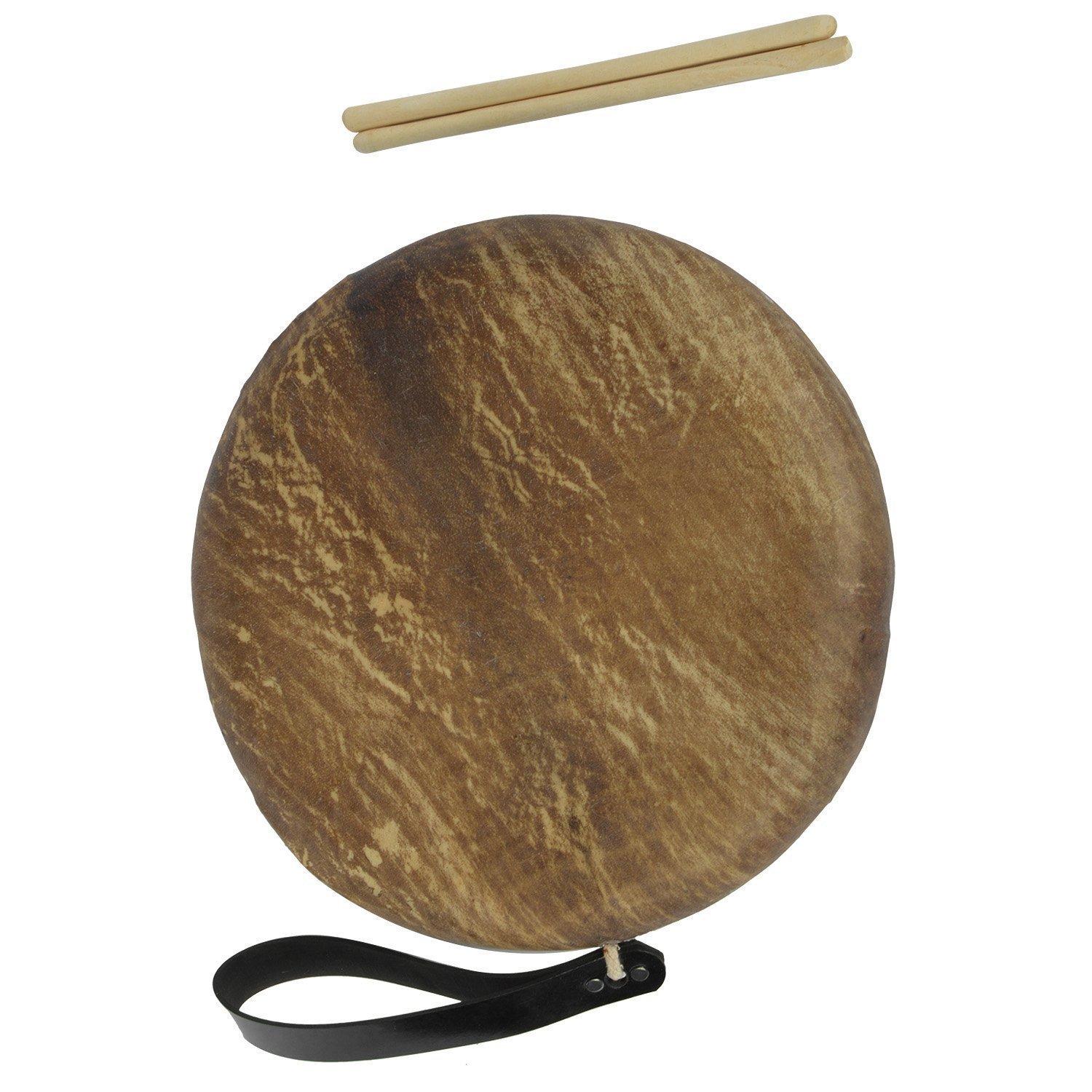} & Q: Is the image about oom-pah? A: No \colorbox{yellow!30}{Parai} \colorbox{mygreen!30}{TA} \colorbox{blue!30}{Speech and Language} \\\midrule
\includegraphics[width=0.9\linewidth]{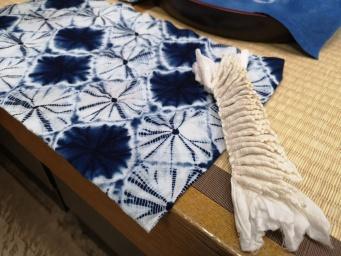} & \colorbox{red!30}{Shibori} is a Japanese manual tie-dyeing technique, which produces a number of different patterns on fabric. & \includegraphics[width=0.9\linewidth]{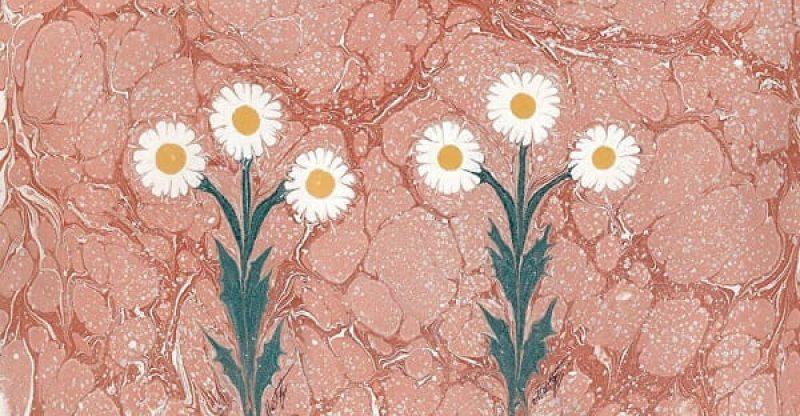} & 
Q: Is this an example of paper marbling? A: No \colorbox{yellow!30}{Paper marbling} \colorbox{mygreen!30}{TR} \colorbox{blue!30}{Speech and Language} \\\midrule
\includegraphics[width=0.9\linewidth]{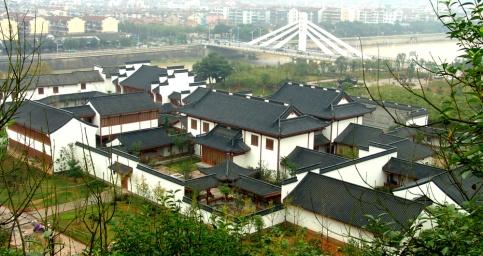} &
\small{A \colorbox{red!30}{siheyuan} is a historical type of residence that was commonly found throughout China, most famously in Beijing and rural Shanxi... remaining siheyuan are often still used as subdivided housing complexes, although many lack modern amenities.} & \includegraphics[width=0.9\linewidth]{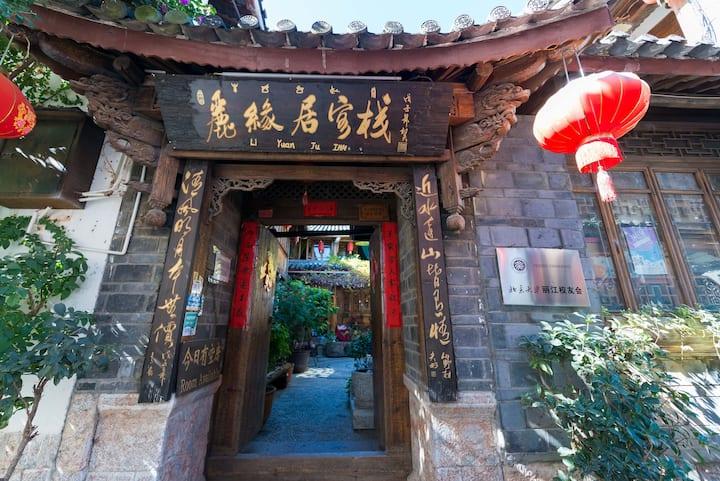} & 
Q: Is there a siheyuan in the image? A: Yes \colorbox{yellow!30}{Siheyuan} \colorbox{mygreen!30}{ZH} \colorbox{blue!30}{The house} \\
      \bottomrule
      \end{tabular}}
      \caption{Dataset examples: \colorbox{red!30}{Question Concept}, \colorbox{yellow!30}{Target Concept}, \colorbox{mygreen!30}{Source Language of Target Concept}, \colorbox{blue!30}{Chapter}. Example 1 (Noken) shows an instance where the textual context complements the image by specifying how the bag is usually worn. Examples 2 and 4 show instances where either the image or the text would be sufficient to answer the question. Example 3 shows an instance where the image exemplar is not as informative, but the text clarifies the type of instrument. Example 5 shows an instance where the text alone does not provide sufficient visual cues.}
      \label{tab:examples}
  \end{table*}

\section{Acknowledgements}
We acknowledge the use of GitHub Copilot\footnote{\href{https://github.com/features/copilot}{https://github.com/features/copilot}} in the implementation of our research.  All final code is verified by the authors.
We also acknowledge the use of ChatGPT\footnote{\href{https://chatgpt.com/}{https://chatgpt.com/}} in improving the clarity of the writing of this paper.

\end{document}